\pdfoutput=1
\documentclass[twoside,11pt]{article}

\usepackage{jmlr2e}
\usepackage{amsmath}
\usepackage{amsfonts}
\usepackage{amssymb}
\usepackage{color,soul}
\usepackage{footnote}
\usepackage{hyperref}
\usepackage{listings}
\usepackage{algpseudocode}
\usepackage{algorithm}
\usepackage{float}
\usepackage{array}
\usepackage{setspace}
\newcolumntype{H}{>{\setbox0=\hbox\bgroup}c<{\egroup}@{}}

\DeclareMathOperator*{\argmax}{arg\,max}

\ShortHeadings{Benchmarking for Bayesian Reinforcement Learning}{Castronovo and Ernst and Co\"etoux and Fonteneau}
\firstpageno{1}

\begin{document}

\title{Benchmarking for Bayesian Reinforcement Learning}

\author{\name Micha\"el Castronovo \email m.castronovo@ulg.ac.be \\
		\addr Montefiore Institute \\
		University of Li\`ege \\
		B-4000 Li\`ege, Belgium
		\AND
		\name Damien Ernst \email dernst@ulg.ac.be \\
		\addr Montefiore Institute \\
		University of Li\`ege \\
		B-4000 Li\`ege, Belgium
		\AND
		\name Adrien Cou\"etoux \email acouetoux@ulg.ac.be \\
		\addr Montefiore Institute \\
		University of Li\`ege \\
		B-4000 Li\`ege, Belgium	
		\AND
		\name Rapha\"el Fonteneau \email raphael.fonteneau@ulg.ac.be \\
		\addr Montefiore Institute \\
		University of Li\`ege \\
		B-4000 Li\`ege, Belgium
}


\maketitle

\begin{abstract}
In the Bayesian Reinforcement Learning (BRL) setting, agents try to maximise the collected rewards while interacting with their environment while using some prior knowledge that is accessed beforehand. Many BRL algorithms have already been proposed, but even though a few toy examples exist in the literature, there are still no extensive or rigorous benchmarks to compare them.
The paper addresses this problem, and provides a new BRL comparison methodology along with the corresponding open source library. In this methodology, a comparison criterion that measures the performance of algorithms on large sets of Markov Decision Processes (MDPs) drawn from some probability distributions is defined.
In order to enable the comparison of non-anytime algorithms, our methodology also includes a detailed analysis of the computation time requirement of each algorithm.
Our library is released with all source code and documentation: it includes three test problems, each of which has two different prior distributions, and seven state-of-the-art RL algorithms.
Finally, our library is illustrated by comparing all the available algorithms and the results are discussed.
\end{abstract}


\section{Introduction}
Reinforcement Learning (RL) agents aim to maximise collected rewards by interacting over a certain period of time in initially unknown environments. Actions that yield the highest performance according to the current knowledge of the environment and those that maximise the gathering of new knowledge on the environment may not be the same. This is the dilemma known as Exploration/Exploitation (E/E). In such a context, using prior knowledge of the environment is extremely valuable, since it can help guide the decision-making process in order to reduce the time spent on exploration. Model-based Bayesian Reinforcement Learning (BRL) (\cite{Dearden99, Strens00}) specifically targets RL problems for which such a prior knowledge is encoded in the form of a probability distribution (the ``prior'') over possible models of the environment. As the agent interacts with the actual model, this probability distribution is updated according to the Bayes rule into what is known as ``posterior distribution''. The BRL process may be divided into two learning phases: the offline learning phase refers to the phase when the prior knowledge is used to warm-up the agent for its future interactions with the real model. The online learning phase, on the other hand, refers to the actual interactions between the agent and the model. In many applications, interacting with the actual environment may be very costly (e.g. medical experiments). In such cases, the experiments made during the online learning phase are likely to be much more expensive than those performed during the offline learning phase.

In this paper, we investigate how the way BRL algorithms use the offline learning phase may impact online performances. To properly compare Bayesian algorithms, the first comprehensive BRL benchmarking protocol is designed, following the foundations of~\cite{Castronovo14}. ``Comprehensive BRL benchmark'' refers to a tool which assesses the performance of BRL algorithms over a large set of problems that are actually drawn according to a prior distribution. In previous papers addressing BRL, authors usually validate their algorithm by testing it on a few test problems, defined by a small set of predefined MDPs. For instance, BAMCP (\cite{Guez2012}), SBOSS (\cite{Castro10}), and BFS3 (\cite{Asmuth11approachingbayes-optimalilty}) are all validated on a fixed number of MDPs. In their validation process, the authors select a few BRL tasks, for which they choose one arbitrary transition function, which defines the corresponding MDP. Then, they define one prior distribution compliant with the transition function. This type of benchmarking is problematic in the sense that the authors actually know the hidden transition function of each test case. 
It also creates an implicit incentive to over-fit their approach to a few specific transition functions, which should be completely unknown before interacting with the model. In this paper, we compare BRL algorithms in several different tasks. In each task, the real transition function is defined using a random distribution, instead of being arbitrarily fixed. Each algorithm is thus tested on an infinitely large number of MDPs, for \emph{each} test case. To perform our experiments, we developed the \textit{BBRL} library, whose objective is to also provide other researchers with our benchmarking tool. 

This paper is organised as follows:
Section~\ref{problem_statement} presents the problem statement.
Section~\ref{bayesian_rl_protocol} formally defines the experimental protocol designed for this paper.
Section~\ref{bbrl_library} briefly presents the library.
Section~\ref{application} shows a detailed application of our protocol, comparing several well-know BRL algorithms on three different benchmarks.
Section~\ref{conclusion} concludes the study.

\section{Problem Statement}
\label{problem_statement}

This section is dedicated to the formalisation of the different tools and concepts discussed in this paper.

\subsection{Reinforcement Learning}
\label{problem_statement:reinforcement_learning}
Let $M = (X, U, f(\cdot), \rho_M, p_{M, 0}(\cdot), \gamma)$ be a given unknown MDP, where $X = \{ x^{(1)}, \ldots,  x^{(n_X)} \}$ denotes its finite state space and $U = \{u^{(1)}, \ldots,  u^{(n_U)} \}$ refers to its finite action space. 
When the MDP is in state $x_t$ at time $t$ and action $u_t$ is selected, the agent moves instantaneously to a next state $x_{t + 1}$ with  a probability of $P(x_{t+1} | x_t, u_t) = f(x_t, u_t, x_{t + 1})$. 
An instantaneous deterministic, bounded reward $r_t = \rho_M(x_t, u_t, x_{t+1}) \in [ R_{\min}, R_{\max}  ]$ is observed. 

Let  $h_t = ( x_0, u_0, r_0, x_1, \cdots, x_{t - 1}, u_{t - 1}, r_{t -1}, x_t) \in H$ denote the history observed until time $t$. 
An E/E strategy is a stochastic policy $\pi$ which, given the current state $x_t$, returns an action $u_t \sim \pi(h_t)$. 
Given a probability distribution over initial states $p_{M, 0}(\cdot)$, the expected return of a given E/E strategy $\pi$ with respect to the MDP $M$ can be defined as follows:
\begin{equation*}
J^\pi_M = \underset {x_0 \sim p_{M, 0}(\cdot)} {\mathbb{E}} [\mathcal{R}^\pi_M(x_0)],
\end{equation*}
where $\mathcal{R}^\pi_M(x_0)$ is the stochastic sum of discounted rewards received when applying the policy $\pi$, starting from an initial state $x_0$:
\[
	\mathcal{R}^\pi_M(x_0) = \sum_{t = 0}^{+ \infty} \gamma^t \; r_t.
\]
RL aims to learn the behaviour that maximises $J^\pi_M$, i.e. learning a policy $\pi^*$ defined as follows:
\begin{eqnarray*}
\pi^* \in \underset{\pi}{\arg\max} \quad J^\pi_M.
\end{eqnarray*}

\subsection{Prior Knowledge}
\label{problem_statement:prior_distribution}
In this paper, the actual MDP is assumed to be initially unknown. Model-based Bayesian Reinforcement Learning (BRL) proposes to the model the uncertainty, using a probability distribution $p^{0}_{\mathcal M}(\cdot)$ over a set of candidate MDPs $\mathcal M $. Such a probability distribution is called a prior distribution and can be used to encode specific prior knowledge available before interaction. Given a prior distribution $p^{0}_{\mathcal M}(\cdot)$, the expected return of a given E/E strategy $\pi$ is defined as:
\begin{equation*}
\mathfrak J^\pi_{p^{0}_{\mathcal M}(\cdot)} = \underset {M \sim p^{0}_{\mathcal M}(\cdot)} {\mathbb{E}} \left[J^\pi_M\right],
\end{equation*}
In the BRL framework, the goal is to maximise $\mathfrak J^\pi_{p^{0}_{\mathcal M}(\cdot)}$, by finding $\pi^*$, which is called ``Bayesian optimal policy'' and defined as follows:
\begin{eqnarray*}
\pi^* \in \underset{\pi}{\arg\max} \quad \mathfrak  J^\pi_{p^{0}_{\mathcal M}(\cdot)}.
\end{eqnarray*}

\subsection{Computation time characterisation}
\label{problem_statement:time_constraints}
Most BRL algorithms rely on some properties which, given sufficient computation time, ensure that their agents will converge to an optimal behaviour. However, it is not clear to know beforehand whether an algorithm will satisfy fixed computation time constraints while providing good performances.

The parameterisation of the algorithms makes the selection even more complex.
Most BRL algorithms depend on parameters (number of transitions simulated at each iteration, etc.) which, in some way, can affect the computation time.
In addition, for one given algorithm and fixed parameters, the computation time often varies from one simulation to another.
These features make it nearly impossible to compare BRL algorithms under strict computation time constraints.
In this paper, to address this problem, algorithms are run with multiple choices of parameters, and we analyse their time performance a posteriori.

Furthermore, a distinction between the offline and online computation time is made.
Offline computation time corresponds to the moment when the agent is able to exploit its prior knowledge, but cannot interact with the MDP yet.
One can see it as the time given to take the first decision.
In most algorithms concerned in this paper, this phase is generally used to initialise some data structure.
On the other hand, online computation time corresponds to the time consumed by an algorithm for taking each decision.

There are many ways to characterise algorithms based on their computation time.
One can compare them based on the average time needed per step or on the offline computation time alone.
To remain flexible, for each run of each algorithm, we store its computation times $(B_i)_{-1 \leq i}$, with $i$ indexing the time step, and $B_{-1}$ the offline learning time.
Then a feature function $\phi((B_i)_{-1 \leq i})$ is extracted from this data.
This function is used as a metric to characterise and discriminate algorithms based on their time requirements.

In our protocol, which is detailed in the next section, two types of characterisation are used. 
For a set of experiments, algorithms are classified based on their offline computation time only, i.e. we use $\phi((B_i)_{-1 \leq i}) =  B_{-1}$.
Afterwards, the constraint is defined as $\phi((B_i)_{-1 \leq i}) \leq K$, $K>0$ in case it is required to only compare the algorithms that have an offline computation time lower than $K$.

For another set of experiments, algorithms are separated according to their empirical average online computation time.
In this case, $\phi((B_i)_{-1 \leq i}) = \frac{1}{n}\sum_{0 \leq i < n} B_i$.
Algorithms can then be classified based on whether or not they respect the constraint $\phi((B_i)_{-1 \leq i}) \leq K$, $K>0$.

This formalisation could be used for any other computation time characterisation.
For example, one could want to analyse algorithms based on the longest computation time of a trajectory, and define $\phi((B_i)_{-1 \leq i}) = \max_{-1 \leq i} B_i $.

\section{A new Bayesian Reinforcement Learning benchmark protocol}
\label{bayesian_rl_protocol}

\subsection{A comparison criterion for BRL}
In this paper, a real Bayesian evaluation is proposed, in the sense that the different algorithms are compared on a large set of problems drawn according to a test probability distribution.
This is in contrast with the Bayesian literature (\cite{Guez2012,Castro10,Asmuth11approachingbayes-optimalilty}), where authors pick a fixed number of MDPs on which they evaluate their algorithm.

Our criterion to compare algorithms is to measure their average rewards against a given random distribution of MDPs, using another distribution of MDPs as a prior knowledge.
In our experimental protocol, an experiment is defined by a prior distribution $p^{0}_{\mathcal M}(\cdot)$ and a test distribution $p_{\mathcal M}(\cdot)$. 
Both are random distributions over the set of possible MDPs, not stochastic transition functions.
To illustrate the difference, let us take an example.
Let $(x,u,x')$ be a transition.
Given a transition function $f : X \times U \times X \rightarrow [0; 1]$, $f(x,u,x')$ is the probability of observing $x'$ if we chose $u$ in $x$.
In this paper, this function $f$ is assumed to be the only unknown part of the MDP that the agent faces.
Given a certain test case, $f$ corresponds to a unique MDP $M \in \mathcal{M}$.
A Bayesian learning problem is then defined by a probability distribution over a set $\mathcal{M}$ of possible MDPs.
We call it a test distribution, and denote it $p_{\mathcal M}(\cdot)$.
Prior knowledge can then be encoded as another distribution over $\mathcal{M}$, and denoted $p^{0}_{\mathcal M}(\cdot)$.
We call ``accurate'' a prior which is identical to the test distribution ($p^{0}_{\mathcal M}(\cdot) = p_{\mathcal M}(\cdot)$), and we call ``inaccurate'' a prior which is different ($p^{0}_{\mathcal M}(\cdot) \neq p_{\mathcal M}(\cdot)$).

In previous Bayesian literature, authors select a fixed number of MDPs $M_1, ..., M_n$, train and test their algorithm on them.
Doing so does not guarantee any generalisation capabilities.
To solve this problem, a protocol that allows rigorous comparison of BRL algorithms is designed.
Training and test data are separated, and can even be generated from different distributions (in what we call the inaccurate case).
 
More precisely, our protocol can be described as follows:
Each algorithm is first trained on the prior distribution. 
Then, their performances are evaluated by estimating the expectation of the discounted sum of rewards, when they are facing MDPs drawn from the test distribution. 
Let $\mathfrak J^{\pi(p^{0}_{\mathcal{M}})}_{p_{\mathcal{M}}}$ be this value:
\[
\mathfrak	J^{\pi(p^{0}_{\mathcal{M}})}_{p_{\mathcal{M}}} = \underset {M \sim p_{\mathcal{M}}} {\mathbb{E}} \left[ J^{\pi(p^{0}_{\mathcal{M}})}_M \right],
\]
where $\pi(p^0_{\mathcal{M}})$ is the algorithm $\pi$ trained offline on $p^0_{\mathcal{M}}$. In our Bayesian RL setting, we want to find the algorithm $\pi^*$ which maximises $\mathfrak J^{\pi(p^{0}_{\mathcal{M}})}_{p_{\mathcal{M}}}$ for the $\langle p^0_{\mathcal{M}}, p_{\mathcal{M}} \rangle$ experiment:
\[
	\pi^* \in \underset{\pi}{\arg\max} \quad \mathfrak J^{\pi(p^{0}_{\mathcal{M}})}_{p_{\mathcal{M}}}.
\]

In addition to the performance criterion, we also measure the empirical computation time.
In practice, all problems are subject to time constraints.
Hence, it is important to take this parameter into account when comparing different algorithms.

\subsection{The experimental protocol}
In practice, we can only sample a finite number of trajectories, and must rely on estimators to compare algorithms.
In this section our experimental protocol is described, which is based on our comparison criterion for BRL and provides a detailed computation time analysis.

An experiment is defined by (i) a prior distribution $p^{0}_{\mathcal{M}}$ and (ii) a test distribution $p_{\mathcal{M}}$.
Given these, an agent is evaluated $\pi$ as follows:

\begin{enumerate}
	\item Train $\pi$ offline on $p^{0}_{\mathcal{M}}$.
	\item Sample $N$ MDPs from the test distribution $p_{\mathcal{M}}$.
	\item For each sampled MDP $M$, compute estimate $\bar{J}^{\pi(p^{0}_{\mathcal{M}})}_{M}$ of $J^{\pi(p^{0}_{\mathcal{M}})}_M$.
	\item Use these values to compute an estimate $\bar{\mathfrak J}^{\pi(p^{0}_{\mathcal{M}})}_{p_{\mathcal{M}}}$.
\end{enumerate}

To estimate $J^{\pi(p^{0}_{\mathcal{M}})}_M$, the expected return of agent $\pi$ trained offline on $p^{0}_{\mathcal{M}}$, one trajectory is sampled on the MDP $M$, and the cumulated return is computed $\bar{J}^{\pi(p^{0}_{\mathcal{M}})}_{M_i} = \mathcal{R}_{M}^{\pi(p_{\mathcal{M}}^0)}(x_0)$.

To estimate this return, each trajectory is truncated after $T$ steps. 
Therefore, given an MDP $M$ and its initial state $x_0$, we observe $\bar{\mathcal{R}}^{\pi(p^{0}_{\mathcal{M}})}_M(x_0)$, an approximation of $\mathcal{R}^{\pi(p^{0}_{\mathcal{M}})}_M(x_0)$:
\[
\bar{\mathcal{R}}^{\pi(p^{0}_{\mathcal{M}})}_M(x_0) = \sum_{t = 0}^{T} \gamma^t r_t.
\]

If $R_{max}$ denotes the maximal instantaneous reward an agent can receive when interacting with an MDP drawn from $p_{\mathcal M}$, then choosing $T$ as guarantees the approximation error is bounded by $\epsilon >0$:
\[
T = \left \lfloor \frac{\log(\epsilon \times \frac{ (1 - \gamma)}{R_{max}} )}{\log{\gamma}} \right \rfloor.
\]
$\epsilon = 0.01$ is set for all experiments, as a compromise between measurement accuracy and computation time.

Finally, to estimate our comparison criterion $\mathfrak J^{\pi(p^{0}_{\mathcal{M}})}_{p_{\mathcal{M}}}$, the empirical average of the algorithm performance is computed over $N$ different MDPs, sampled from $p_{\mathcal{M}}$ :
\begin{equation}
\bar{\mathfrak J}^{\pi(p^{0}_{\mathcal{M}})}_{p_{\mathcal{M}}} = \frac{1}{N}\sum_{0\leq i < N} \bar{J}^{\pi(p^{0}_{\mathcal{M}})}_{M_i} =  \frac{1}{N}\sum_{0\leq i < N}\bar{\mathcal{R}}^{\pi(p^{0}_{\mathcal{M}})}_{M_i}(x_0)
\end{equation}\\

For each agent $\pi$, we retrieve $\mu_{\pi}=\bar{J}^{\pi}_M$ and $\sigma_{\pi}$, the empirical mean and standard deviation of the results observed respectively. 
This gives us the following statistical confidence interval at 95\% for $J^\pi_M$:
\[
	J^\pi_M \in \left[ \bar{J}^{\pi}_M - \frac{2\sigma_{\pi}}{N}; \; \bar{J}^{\pi}_M + \frac{2\sigma_{\pi}}{N} \right].
\]
The values reported in the following figures and tables are estimations of the interval within which $J^\pi_M$ is, with probability $0.95$.\\

As introduced in Section \ref{problem_statement:time_constraints}, in our methodology, a function $\phi$ of computation times is used to classify algorithms based on their time performance.
The choice of $\phi$ depends on the type of time constraints that are the most important to the user.
In this paper, we reflect this by showing three different ways to choose $\phi$.
These three choices lead to three different ways to look at the results and compare algorithms.
The first one is to classify algorithms based on their offline computation time, the second one is to classify them based on the algorithms average online computation time. The third is a combination of the first two choices of $\phi$, that we denote $\phi_{off}((B_i)_{-1 \leq i}) = B_{-1}$ and  $\phi_{on}((B_i)_{-1 \leq i}) = \frac{1}{n}\sum_{0 \leq i < n} B_i$.
The objective is that for each pair of constraints $\phi_{off}((B_i)_{-1 \leq i}) < K_1$ and $\phi_{on}((B_i)_{-1 \leq i}) < K_2$, $K_1, K_2 >0$, we want to identify the best algorithms that respect these constraints.
In order to achieve this: (i) All agents that do not satisfy the constraints are discarded; (ii) for each algorithm, the agent leading to the best performance in average is selected; (iii) we build the list of agents whose performances are not significantly different\footnote{A paired sampled $Z$-test with a confidence level of 95\% has been used to determine when two agents are statistically equivalent (more details in Appendix~\ref{appendix:z-test}).}.

The results will help us to identify, for each experiment, the most suitable algorithm(s) depending on the constraints the agents must satisfy.
This protocol is an extension of the one presented in~\cite{Castronovo14}.

\section{BBRL library}
\label{bbrl_library}
\textit{BBRL}\footnote{BBRL stands for \textbf{B}enchmaring tools for \textbf{B}ayesian \textbf{R}einforcement \textbf{L}earning.} is a C++ open-source library for Bayesian Reinforcement Learning (discrete state/action spaces).
This library provides high-level features, while remaining as flexible and documented as possible to address the needs of any researcher of this field.
To this end, we developed a complete command-line interface, along with a comprehensive website:

\begin{center}
	\url{https://github.com/mcastron/BBRL}
\end{center}

\textit{BBRL} focuses on the core operations required to apply the comparison benchmark presented in this paper. To do a complete experiment with the BBRL library, follow these five steps:

\begin{enumerate}
	\item	We create a test and a prior distribution.
	Those distributions are represented by Flat Dirichlet Multinomial distributions (FDM), parameterised by a state space $X$, an action space $U$, a vector of parameters $\boldsymbol \theta$, and reward function $\rho$. For more information about the FDM distributions, check Section~\ref{application:benchmarks}.\\
						\begin{lstlisting}[frame=tlrb,basicstyle=\small\ttfamily]
./BBRL-DDS --mdp_distrib_generation \
              --name <name> \
              --short_name <short name> \
              --n_states <$n_X$> --n_actions <$n_U$> \
              --ini_state <$x_0$> \
              --transition_weights \
                <$\theta(1)$> $\cdots$ <$\theta(n_X n_U n_X)$> \
              --reward_type "RT_CONSTANT" \
              --reward_means \
                <$\rho(x^{(1)}, u^{(1)}, x^{(1)})$> $\cdots$ <$\rho(x^{(n_X)}, u^{(n_U)}, x^{(n_X)})$> \
              --output <output file>
	\end{lstlisting}
A distribution file is created.\\

	\item	We create an experiment.
	An experiment is defined by a set of $N$ MDPs, drawn from a test distribution defined in a \textit{distribution file}, a discount factor $\gamma$ and a horizon limit $T$.
				\begin{lstlisting}[frame=tlrb,basicstyle=\small\ttfamily]
./BBRL-DDS --new_experiment \
              --name <name> \
              --mdp_distribution "DirMultiDistribution" \
                --mdp_distribution_file <distribution file> \
              --n_mdps <$N$> --n_simulations_per_mdp 1 \
              --discount_factor <$\gamma$> --horizon_limit <$T$> \
              --compress_output \
              --output <output file>
	\end{lstlisting}
	An experiment file is created and can be used to conduct the same experiment for several agents.\\

	\item	We create an agent.
	An agent is defined by an algorithm $alg$, a set of parameters $\psi$, and a prior distribution defined in a \textit{distribution file}, on which the created agent will be trained.
				\begin{lstlisting}[frame=tlrb,basicstyle=\small\ttfamily]
./BBRL-DDS --offline_learning \
              --agent <$alg$> [<parameters $\psi$>]\
              --mdp_distribution "DirMultiDistribution" \
                --mdp_distribution_file <distribution file> \
              --output <output file>
	\end{lstlisting}
	An agent file is created.
	The file also stores the computation time observed during the offline training phase.\\

	\item	We run the experiment.
	We need to provide an \textit{experiment file}, an algorithm $alg$ and an \textit{agent file}.
				\begin{lstlisting}[frame=tlrb,basicstyle=\small\ttfamily]
./BBRL-DDS --run_experiment \
              --experiment \
                --experiment_file <experiment file> \
              --agent <$alg$> \
                --agent_file <agent file> \
              --n_threads 1 \
              --compress_output \
              --safe_simulations \
              --refresh_frequency 60 \
              --backup_frequency 900 \
              --output <output file>
	\end{lstlisting}
	A \textit{result file} is created.
	This file contains a set of all transitions encountered during each trajectory.
	Additionally, the computation times we observed are also stored in this file.
	It is often impossible to measure precisely the computation time of a single decision.
	This is why only the computation time of each trajectory is reported in this file.\\

	\item	Our results are exported.
	After each experiment has been performed, a set of $K$ \textit{result files} is obtained.
	We need to provide all \textit{agent files} and \textit{result files} to export the data.
				\begin{lstlisting}[frame=tlrb,basicstyle=\small\ttfamily]
./BBRL-export --agent <$alg^{(1)}$> \
                   --agent_file <agent file #1> \
                 --experiment \
                   --experiment_file <result file #1> \
                 ...
                 --agent <$alg^{(K)}$> \
                   --agent_file <agent file #K> \
                 --experiment \
                   --experiment_file <result file #K>
	\end{lstlisting}
\textit{BBRL} will sort the data automatically and produce several files for each experiment.
	\begin{itemize}
		\item	A graph comparing offline computation cost w.r.t. performance;
		\item	A graph comparing online computation cost w.r.t. performance;
		\item	A graph where the X-axis represents the offline time bound, while the Y-axis represents the online time bound.
		A point of the space corresponds to set of bounds.
		An algorithm is associated to a point of the space if its best agent, satisfying the constraints, is among the best ones when compared to the others;
		\item	A table reporting the results of each agent.
	\end{itemize}
	\textit{BBRL} will also produce a report file in \LaTeX\ gathering the 3 graphs and the table for each experiment.
\end{enumerate}

More than $2.000$ commands have to be entered in order to reproduce the results of this paper.
We decided to provide several $Lua$ script in order to simplify the process.
By completing some configuration files, the user can define the agents, the possible values of their parameters and the experiments to conduct.

\begin{figure}[H]
\begin{lstlisting}[frame=tlrb,basicstyle=\small\ttfamily]
local agents =
{
   --   e-Greedy
   {
      name = "EGreedyAgent",
      params =
      {
         {
            opt = "--epsilon",
            values =
            {
               0.0, 0.1, 0.2, 0.3, 0.4, 0.5, 0.6, 0.7, 0.8, 0.9, 1.0
            }
         }
      },
      olOptions = { "--compress_output" },
      memory = { ol = "1000M", re = "1000M" },
      duration = { ol = "01:00:00", re = "01:00:00" }
   },
   ...
}
\end{lstlisting}
\caption{Example of a configuration file for the agents.}
\end{figure}

\begin{figure}[H]
\begin{lstlisting}[frame=tlrb,basicstyle=\small\ttfamily]
local experiments =
{
   {
      prior = "GC",   priorFile = "GC-distrib.dat",
      exp   = "GC",   testFile  = "GC-distrib.dat",
      N = 500, gamma = 0.95, T = 250
   },
   ...
}
\end{lstlisting}
\caption{Example of a configuration file for the experiments.}
\end{figure}

Those configuration files are then used by a script called \verb?make_scripts.sh?, included within the library, whose purpose is to generate four other scripts:

\begin{itemize}
	\item	\verb?0-init.sh? \\
	Create the experiment files, and create the formulas sets required by OPPS agents.
	\item	\verb?1-ol.sh? \\
	Create the agents and train them on the prior distribution(s).
	\item	\verb?2-re.sh? \\
	Run all the experiments.
	\item	\verb?3-export.sh? \\
	Generate the \LaTeX\ reports.
\end{itemize}

Due to the high computation power required, we made those scripts compatible with workload managers such as SLURM.
In this case, each cluster should provide the same amount of CPU power in order to get consistent time measurements.
To sum up, when the configuration files are completed correctly, one can start the whole process by executing the four scripts, and retrieve the results in nice \LaTeX\ reports.

It is worth noting that there is no computation budget given to the agents.
This is due to the diversity of the algorithms implemented.
No algorithm is ``anytime'' natively, in the sense that we cannot stop the computation at any time and receive an answer from the agent instantly.
Strictly speaking, it is possible to develop an anytime version of some of the algorithms considered in \textit{BBRL}.
However, we made the choice to stay as close as possible to the original algorithms proposed in their respective papers for reasons of fairness.
In consequence, although computation time is a central parameter in our problem statement, it is never explicitly given to the agents.
We instead let each agent run as long as necessary and analyse the time elapsed afterwards.

Another point which needs to be discussed is the impact of the implementation of an algorithm on the comparison results.
For each algorithm, many implementations are possible, some being better than others.
Even though we did our best to provide the best possible implementations, \textit{BBRL} does not compare algorithms but rather the implementations of each algorithms.
Note that this issue mainly concerns small problems, since the complexity of the algorithms is preserved.

\section{Illustration}
\label{application}
This section presents an illustration of the protocol presented in Section~\ref{bayesian_rl_protocol}.
We first describe the algorithms considered for the comparison in Section~\ref{application:compared_algorithms}, followed by a description of the benchmarks in Section~\ref{application:benchmarks}.
Section~\ref{application:results} shows and analyses the results obtained.

\subsection{Compared algorithms}
\label{application:compared_algorithms}
In this section, we present the list of the algorithms considered in this study.
The pseudo-code of each algorithm can be found in Appendix~\ref{appendix:pseudocode}.
For each algorithm, a list of ``reasonable'' values is provided to test each of their parameters.
When an algorithm has more than one parameter, all possible parameter combinations are tested.

\subsubsection{Random}
At each time-step $t$, the action $u_t$ is drawn uniformly from $U$.

\subsubsection{$\epsilon$-Greedy}
\label{compared_algo:egreedy}
The $\epsilon$-Greedy agent maintains an approximation of the current MDP and computes, at each time-step, its associated Q-function. The selected action is either selected randomly (with a probability of $\epsilon$ ($1 \geq \epsilon \geq 0$), or greedily (with a probability of $1 - \epsilon$) with respect to the approximated model.\\

\textbf{Tested values:} 
\begin{itemize}
	\item $\epsilon \in \{0.0, 0.1, 0.2, 0.3, 0.4, 0.5, 0.6, 0.7, 0.8, 0.9, 1.0\}$. \\
\end{itemize}

\subsubsection{Soft-max}
The Soft-max agent maintains an approximation of the current MDP and computes, at each time-step, its associated Q-function. The selected action is selected randomly, where the probability to draw an action $u$ is proportional to $Q(x_{t}, u)$. The temperature parameter $\tau$ allows to control the impact of the Q-function on these probabilities ($\tau \rightarrow 0^{+}$: greedy selection; $\tau \rightarrow +\infty$: random selection).\\

\textbf{Tested values:}
\begin{itemize}
	\item	$\tau \in \{0.05, 0.10, 0.20, 0.33, 0.50, 1.0, 2.0, 3.0, 5.0, 25.0\}$. \\
\end{itemize}

\subsubsection{OPPS}
Given a prior distribution $p^{0}_{\mathcal M}(.)$ and an E/E strategy space $\mathcal{S}$ (either discrete or continuous), the Offline, Prior-based Policy Search algorithm (OPPS) identifies a strategy $\pi^{*} \in \mathcal{S}$ which maximises the expected discounted sum of returns over MDPs drawn from the prior. 

The OPPS for Discrete Strategy spaces algorithm (OPPS-DS) (\cite{Castronovo12, Castronovo14}) formalises the strategy selection problem as a $k$-armed bandit problem, where $k = |\mathcal{S}|$. Pulling an arm amounts to draw an MDP from $p^{0}_{\mathcal M}(.)$, and play the E/E strategy associated to this arm on it for one single trajectory. The discounted sum of returns observed is the return of this arm. This multi-armed bandit problem has been solved by using the UCB1 algorithm (\cite{Auer2002,Audibert2007}). The time budget is defined by a variable $\beta$, corresponding to the total number of draws performed by the UCB1.

The E/E strategies considered by Castronovo et. al are index-based strategies, where the index is generated by evaluating a small formula. A formula is a mathematical expression, combining specific features (Q-functions of different models) by using standard mathematical operators (addition, subtraction, logarithm, etc.). The discrete E/E strategy space is the set of all formulas which can be built by combining at most $n$ features/operators (such a set is denoted by $\mathbb{F}_n$).

OPPS-DS does not come with any guarantee. However, the UCB1 bandit algorithm used to identify the best E/E strategy within the set of strategies provides statistical guarantees that the best E/E strategies are identified with high probability after a certain budget of experiments.  However, it is not clear that the best strategy of the E/E strategy space considered yields any high-performance strategy regardless the problem.\\

\textbf{Tested values:}
\begin{itemize}
	\item $\mathcal{S} \in \{\mathbb{F}_2, \mathbb{F}_3, \mathbb{F}_4, \mathbb{F}_5, \mathbb{F}_6\}$\footnote{The number of arms $k$ is always equal to the number of strategies in the given set. For your information: $|\mathbb{F}_2| = 12, |\mathbb{F}_3| = 43, |\mathbb{F}_4| = 226, |\mathbb{F}_5| = 1210, |\mathbb{F}_6| = 7407$},
	\item $\beta \in \{50, 500, 1250, 2500, 5000, 10000, 100000, 1000000\}$. \\
\end{itemize}

\subsubsection{BAMCP}
Bayes-adaptive Monte Carlo Planning (BAMCP) (\cite{Guez2012}) is an evolution of the Upper Confidence Tree (UCT) algorithm (\cite{Kocsis2006}), where each transition is sampled according to the history of observed transitions. 
The principle of this algorithm is to adapt the UCT principle for planning in a Bayes-adaptive MDP, also called the belief-augmented MDP, which is an MDP obtained when considering augmented states made of the concatenation of the actual state and the posterior. The BAMCP algorithm is made computationally tractable by using a sparse sampling strategy, which avoids sampling a model from the posterior distribution at every node of the planification tree. Note that the BAMCP also comes with theoretical guarantees of convergence towards Bayesian optimality.

In practice, the BAMCP relies on two parameters: (i) Parameter $K$ which defines the number of nodes created at each time-step, and (ii) Parameter $depth$ which defines the depth of the tree from the root.\\

\textbf{Tested values:}
\begin{itemize}
	\item $K \in \{1, 500, 1250, 2500, 5000, 10000, 25000\}$,
	\item $depth \in \{15, 25, 50\}$. \\
\end{itemize}

\subsubsection{BFS3}
The Bayesian Forward Search Sparse Sampling (BFS3) (\cite{Asmuth11approachingbayes-optimalilty}) is a Bayesian RL algorithm whose principle is to apply the principle of the FSSS (Forward Search Sparse Sampling, see \cite{Kearns2002}) algorithm to belief-augmented MDPs. It first samples one model from the posterior, 
which is then used to sample transitions. The algorithm then relies on lower and upper bounds on the value of each augmented state to prune the search space. The authors also show that BFS3 converges towards Bayes-optimality as the number of samples increases.

In practice, the parameters of BFS3 are used to control how much computational power is allowed. 
The parameter $K$ defines the number of nodes to develop at each time-step, $C$ defines the branching factor of the tree and $depth$ controls its maximal depth.\\

\textbf{Tested values:} 
\begin{itemize}
	\item $K \in \{1, 500, 1250, 2500, 5000, 10000\}$,
	\item $C \in \{2, 5, 10, 15\}$,
	\item $depth \in \{15, 25, 50\}$. \\
\end{itemize}

\subsubsection{SBOSS}
The Smarter Best of Sampled Set (SBOSS) (\cite{Castro10}) is a Bayesian RL algorithm which relies on the assumption that the model is sampled from a Dirichlet distribution.
From this assumption, it derives uncertainty bounds on the value of state action pairs.
It then uses those bounds to decide how many models to sample from the posterior, and how often the posterior should be updated in order to reduce the computational cost of Bayesian updates.
The sampling technique is then used to build a merged MDP, as in~\cite{Asmuth09}, and to derive the corresponding optimal action with respect to that MDP.
In practice, the number of sampled models is determined dynamically with a parameter $\epsilon$. The re-sampling frequency depends on a parameter $\delta$.\\

\textbf{Tested values:}
\begin{itemize}
	\item $\epsilon \in \{1.0, 1e-1, 1e-2, 1e-3, 1e-4, 1e-5, 1e-6\}$,
	\item $\delta \in \{9, 7, 5, 3, 1, 1e-1, 1e-2, 1e-3n 1e-4, 1e-5, 1e-6\}$. \\
\end{itemize}

\subsubsection{BEB}
The Bayesian Exploration Bonus (BEB) (\cite{Kolter09near-bayesianexploration}) is a Bayesian RL algorithm which builds, at each time-step $t$, the expected MDP given the current posterior. Before solving this MDP, it computes a new reward function $\rho^{(t)}_{BEB}(x, u, y) = \rho_M(x, u, y) + \frac{\beta}{c^{(t)}_{<x, u, y>}}$, where $c^{(t)}_{<x, u, y>}$ denotes the number of times transition $<x, u, y>$ has been observed at time-step $t$. This algorithm solves the mean MDP of the current posterior, in which we replaced $\rho_M(\cdot, \cdot, \cdot)$ by $\rho^{(t)}_{BEB}(\cdot, \cdot, \cdot)$, and applies its optimal policy on the current MDP for one step. The bonus $\beta$ is a parameter controlling the E/E balance. BEB comes with theoretical guarantees of convergence towards Bayesian optimality.\\

\textbf{Tested values:}
\begin{itemize}
	\item $\beta \in \{0.25, 0.5, 1, 1.5, 2, 2.5, 3, 4, 8, 16\}$. \\
\end{itemize}

\subsubsection{Computation times variance}
Each algorithm has one or more parameters that can affect the number of sampled transitions from a given state, or the length of each simulation. This, in turn, impacts the computation time requirement at each step. Hence, for some algorithms, no choice of parameters can bring the computation time below or over certain values.
In other words, each algorithm has its own range of computation time.
Note that, for some methods, the computation time is influenced concurrently by several parameters.
We present a qualitative description of how computation time varies as a function of parameters in Table \ref{table_influence_offline_online}.

\begin{savenotes}
\begin{table}[!h]
\small
\centering
\begin{tabular}{l|l|p{7cm}}
	& \textbf{Offline phase duration} & \textbf{Online phase duration} \\
	\hline
	\textbf{Random} & Almost instantaneous. & Almost instantaneous. \\
	\hline	
	$\boldsymbol \epsilon$\textbf{-Greedy}\footnote{If a random decision is chosen, the model is not solved.}	& Almost instantaneous. & Varies in inverse proportion to $\epsilon$.\\
	& & Can vary a lot from one step to another. \\
	\hline	
	\textbf{OPPS-DS} & Varies proportionally to $\beta$. & Varies proportionally to the number of features implied in the selected E/E strategy. \\
	\hline	
	\textbf{BAMCP}\footnote{$K$ defines the number of nodes to develop at each step, and $depth$ defines the maximal depth of the tree.} & Almost instantaneous. & Varies proportionally to $K$ and $depth$. \\
	\hline	
	\textbf{BFS3}\footnote{$K$ defines the number of nodes to develop at each step, $C$ the branching factor of the tree and $depth$ its maximal depth.} & Almost instantaneous. & Varies proportionally to $K$, $C$ and $depth$. \\
	\hline	
	\textbf{SBOSS}\footnote{The number of models sampled is inversely proportional to $\epsilon$, while the frequency at which the models are sampled is inversely proportional to $\delta$. When an MDP has been sufficiently explored, the number of models to sample and the frequency of the sampling will decrease.} & Almost instantaneous. & Varies in inverse proportion to $\epsilon$ and $\delta$. \\
	& & Can vary a lot from one step to another, with a general decreasing tendency. \\
	\hline	
	\textbf{BEB} & Almost instantaneous. & Constant. \\
\end{tabular}
\caption{Influence of the algorithm and their parameters on the offline and online phases duration.}
\label{table_influence_offline_online}
\end{table}
\end{savenotes}

\subsection{Benchmarks}
\label{application:benchmarks}
In our setting, the transition matrix is the only element which differs between two MDPs drawn from the same distribution.
For each $<$ state, action $>$ pair $<x, u>$, we define a Dirichlet distribution, which represents the uncertainty about the transitions occurring from $<x, u>$. 
A Dirichlet distribution is parameterised by a set of concentration parameters $\alpha_{<x,u>}^{(1)}, \cdots, \alpha_{<x,u>}^{(n_X)}$.

We gathered all concentration parameters in a single vector $\boldsymbol \theta$.
Consequently, our MDP distributions are parameterised by $\rho_M$ (the reward function) and several Dirichlet distributions, parameterised by $\boldsymbol \theta$.
Such a distribution is denoted by $p^{\rho_M, \boldsymbol \theta}(\cdot)$. In the Bayesian Reinforcement Learning community, these distributions are referred to as Flat Dirichlet Multinomial distributions (FDMs).\\

We chose to study two different cases:
\begin{itemize}
	\item	Accurate case: the test distribution is fully known ($p^{0}_{\mathcal M}(.) = p_{\mathcal M}(.)$),
	\item	Inaccurate case: the test distribution is unknown ($p^{0}_{\mathcal M}(.) \neq p_{\mathcal M}(.)$). \\
\end{itemize}

In the inaccurate case, we have no assumption on the transition matrix. We represented this lack of knowledge by a uniform FDM distribution, where each transition has been observed one single time ($\boldsymbol \theta = [1, \cdots, 1]$). \\

Sections~\ref{experiments:gc}, ~\ref{experiments:gdl} and ~\ref{experiments:grid} describes the three distributions considered for this study.

\subsubsection{Generalised Chain distribution ($p^{\rho^{GC}, \boldsymbol \theta^{GC}}(\cdot)$)}
\label{experiments:gc}
The Generalised Chain (GC) distribution is inspired from the five-state chain problem ($5$ states, $3$ actions) (\cite{Dearden98}). The agent starts at State $1$, and has to go through State $2$, $3$ and $4$ in order to reach the last state (State $5$), where the best rewards are. The agent has at its disposal $3$ actions. An action can either let the agent move from State $x^{(n)}$ to State $x^{(n + 1)}$ or force it to go back to State $x^{(1)}$. The transition matrix is drawn from a FDM parameterised by $\boldsymbol \theta^{GC}$, and the reward function is denoted by $\rho^{GC}$. More details can be found in Appendix~\ref{appendix:gc}.

\begin{figure}[H]
	\centering
	\includegraphics[width=0.4\textwidth]{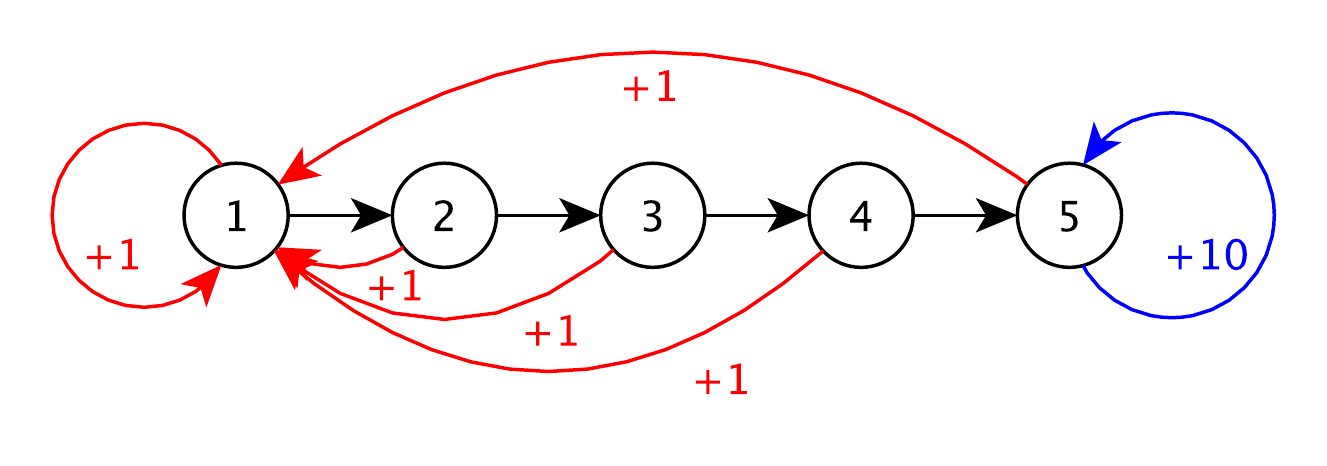}
	\caption{Illustration of the GC distribution.}
\end{figure}

\subsubsection{Generalised Double-Loop distribution ($p^{\rho^{GDL}, \boldsymbol \theta^{GDL}}(\cdot)$)}
\label{experiments:gdl}
The Generalised Double-Loop (GDL) distribution is inspired from the double-loop problem ($9$ states, $2$ actions) (\cite{Dearden98}). Two loops of $5$ states are crossing at State $1$, where the agent starts. One loop is a trap: if the agent enters it, it has no choice to exit but crossing over all the states composing it. Exiting this loop provides a small reward. The other loop is yielding a good reward. However, each action of this loop can either let the agent move to the next state of the loop or force it to return to State $1$ with no reward. The transition matrix is drawn from an FDM parameterised by $\boldsymbol \theta^{GDL}$, and the reward function is denoted by $\rho^{GDL}$. More details can be found in Appendix~\ref{appendix:gdl}.

\begin{figure}[H]
	\centering
	\includegraphics[width=0.4\textwidth]{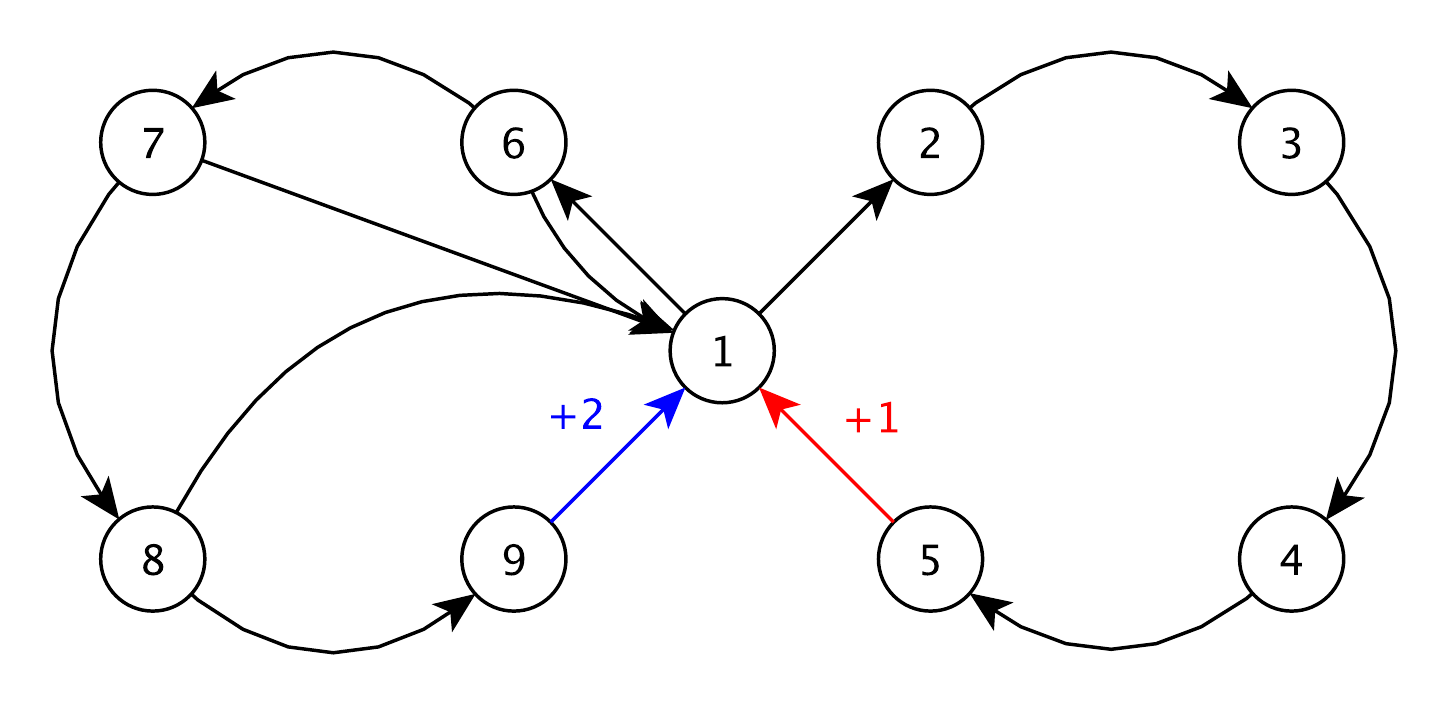}
	\caption{Illustration of the GDL distribution.}
\end{figure}

\subsubsection{Grid distribution ($p^{\rho^{Grid}, \boldsymbol \theta^{Grid}}(\cdot)$)}
\label{experiments:grid}
The Grid distribution is inspired from the Dearden's maze problem ($25$ states, $4$ actions) (\cite{Dearden98}). The agent is placed at a corner of a 5x5 grid (the \textbf{S} cell), and has to reach the opposite corner (the \textbf{G} cell). When it succeeds, it returns to its initial state and receives a reward. The agent can perform $4$ different actions, corresponding to the $4$ directions (up, down, left, right). However, depending on the cell on which the agent is, each action has a certain probability to fail, and can prevent the agent to move in the selected direction. The transition matrix is drawn from an FDM parameterised by $\boldsymbol \theta^{Grid}$, and the reward function is denoted by $\rho^{Grid}$. More details can be found in Appendix~\ref{appendix:grid}.

\begin{figure}[H]
	\centering
	\includegraphics[width=0.35\textwidth]{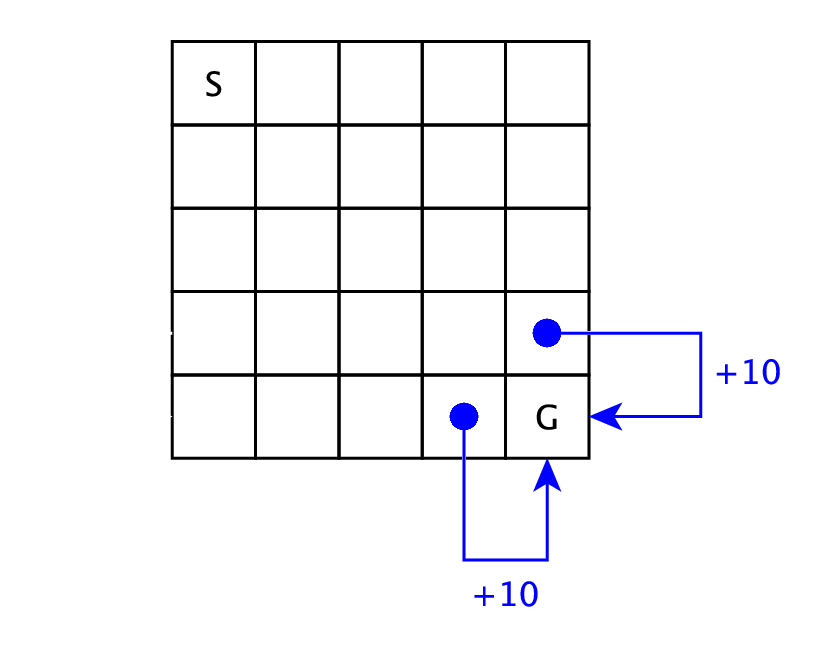}
	\caption{Illustration of the Grid distribution.}
\end{figure}

\newpage
\subsection{Discussion of the results}
\label{application:results}

\subsubsection{Accurate case}
\begin{figure}[H]
\centering
\begin{minipage}{0.45\textwidth}
\centering
\begin{figure}[H]
\centering
\includegraphics[width=\textwidth]{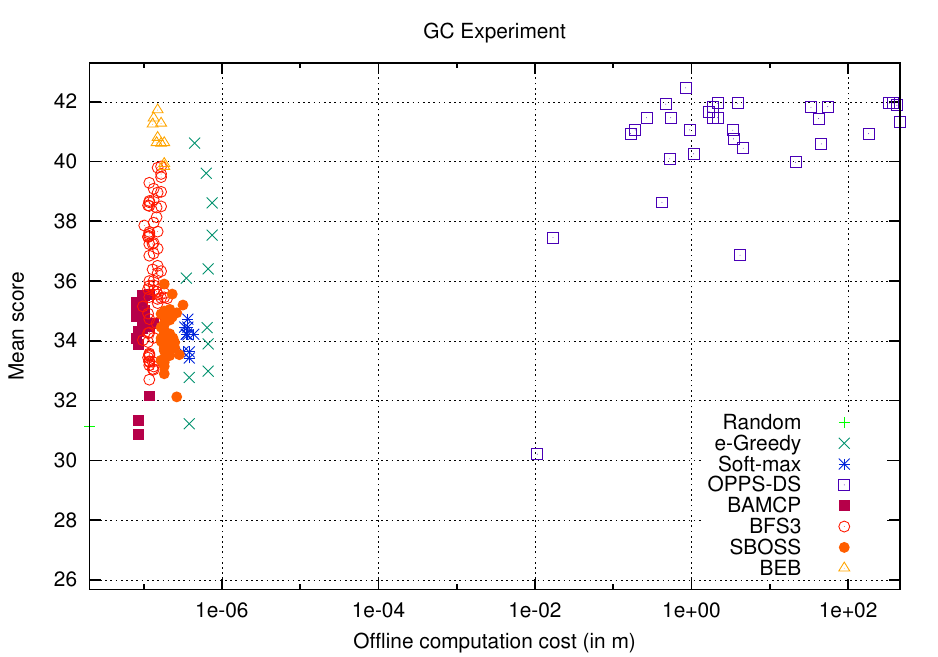}\\
\includegraphics[width=\textwidth]{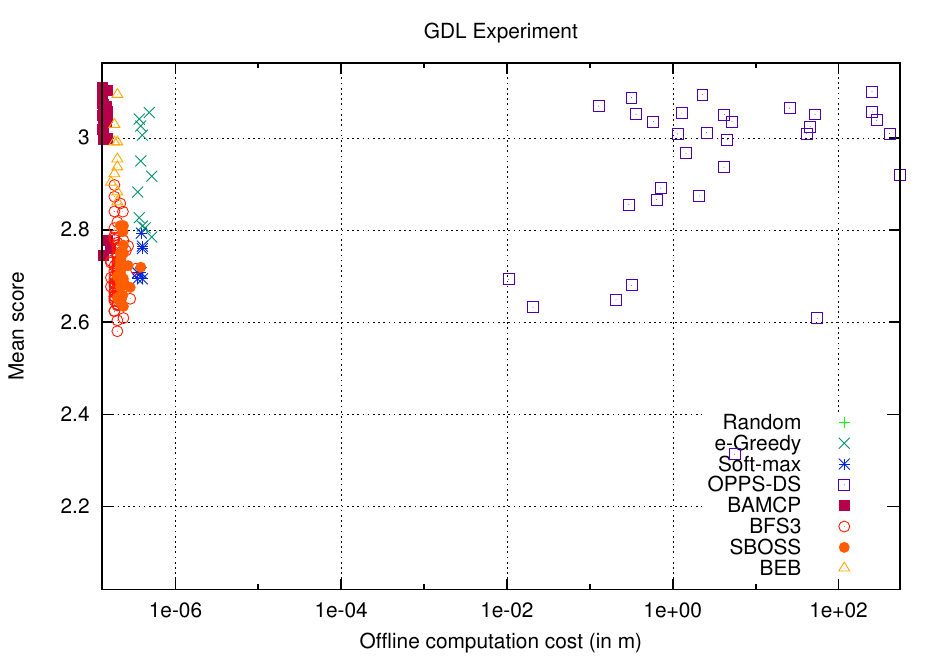}\\
\includegraphics[width=\textwidth]{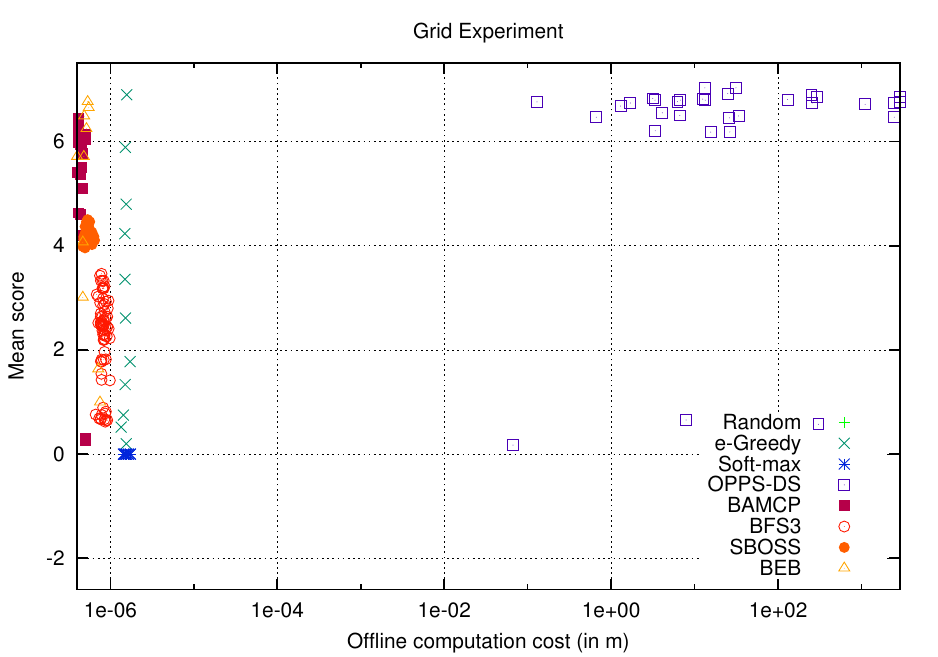}
\end{figure}
\caption{Offline computation cost Vs. Performance}
\label{exp_offline_accurate}
\end{minipage}\hfill
\begin{minipage}{0.45\textwidth}
\centering
\begin{figure}[H]
\centering
\includegraphics[width=\textwidth]{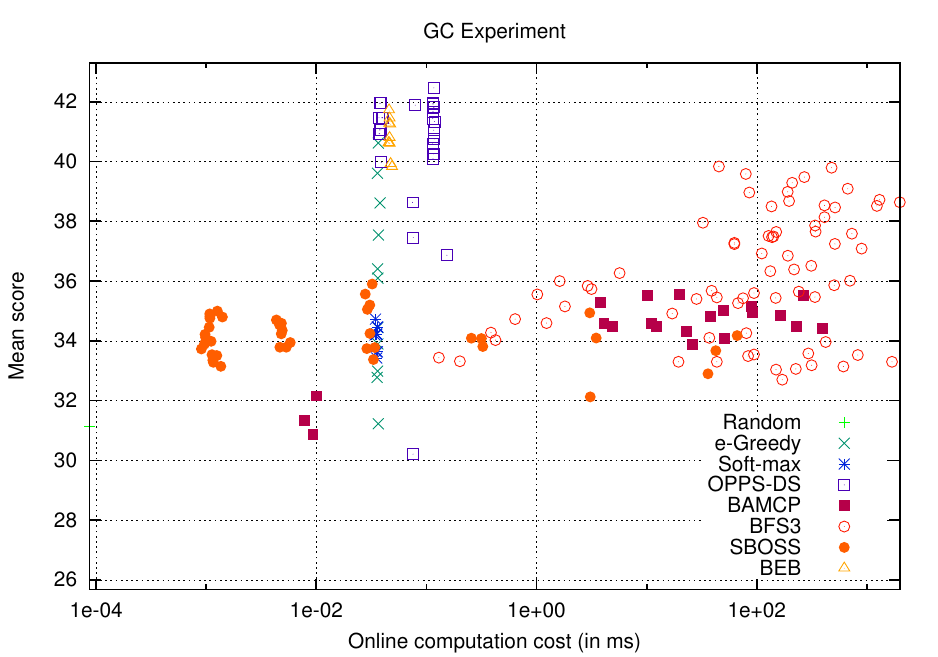}\\
\includegraphics[width=\textwidth]{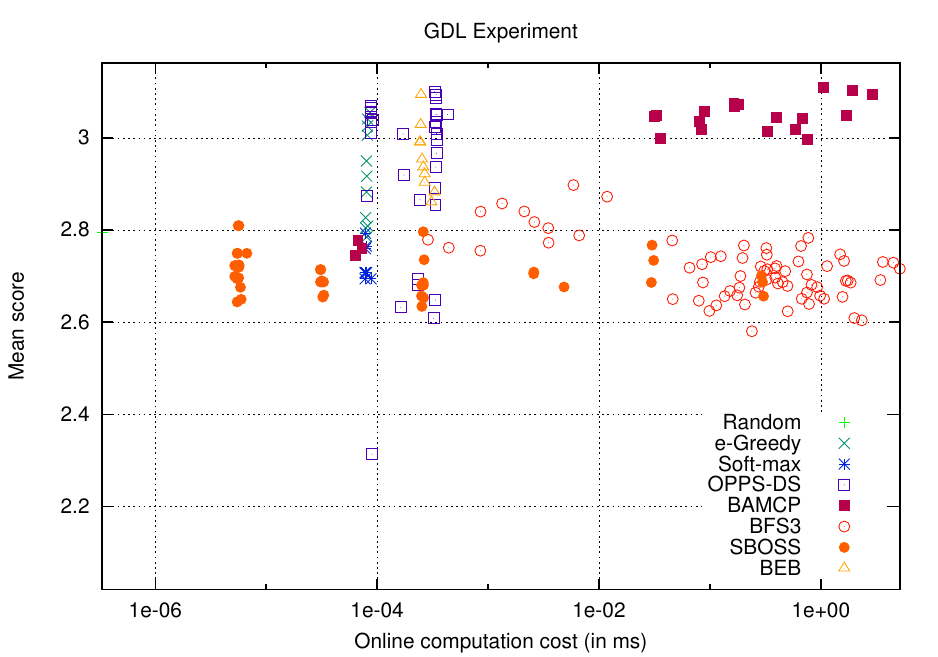}\\
\includegraphics[width=\textwidth]{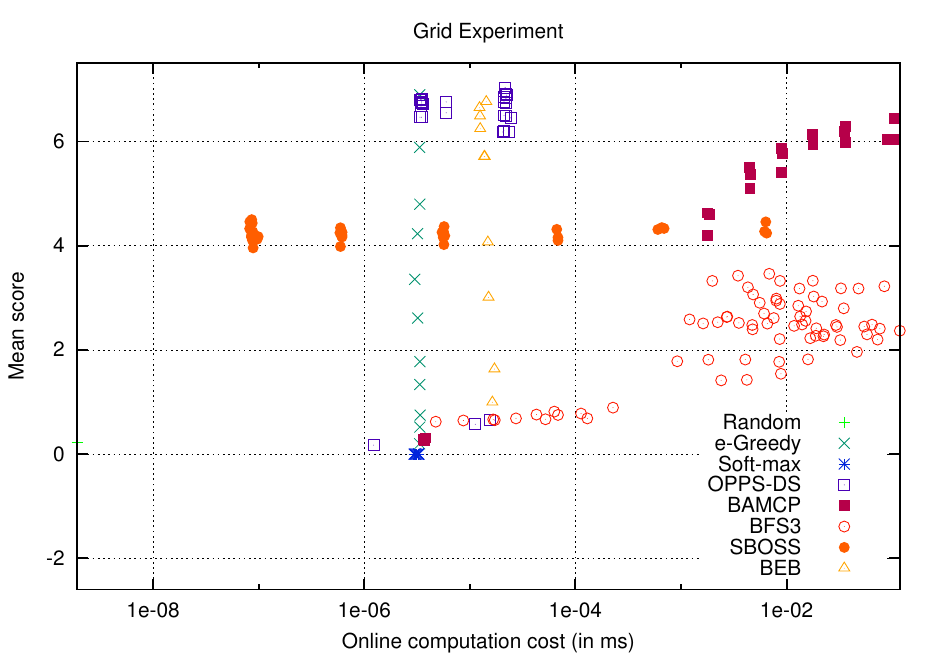}
\end{figure}
\caption{Online computation cost Vs. Performance}
\label{exp_online_accurate}
\end{minipage}
\end{figure}

\newpage
\begin{figure}[H]
\centering
\begin{minipage}{0.45\textwidth}
\centering
\begin{figure}[H]
\centering
\includegraphics[height=6cm]{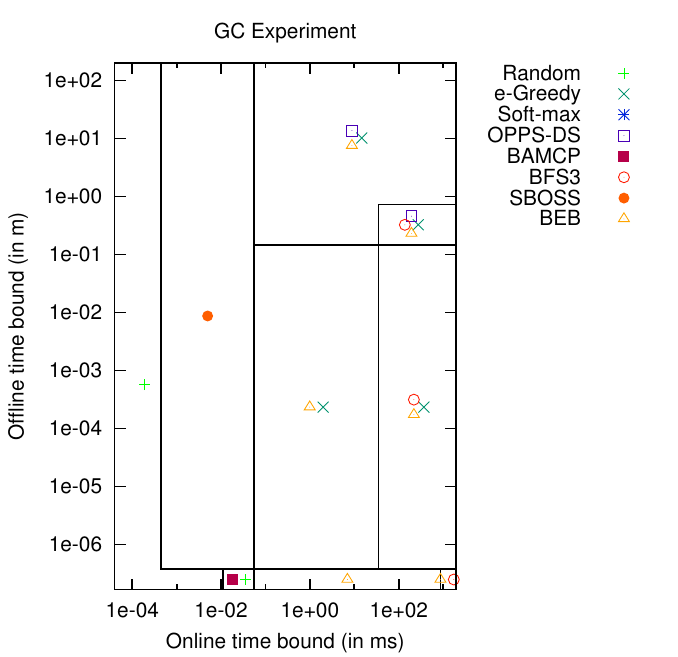}\\
\includegraphics[height=6cm]{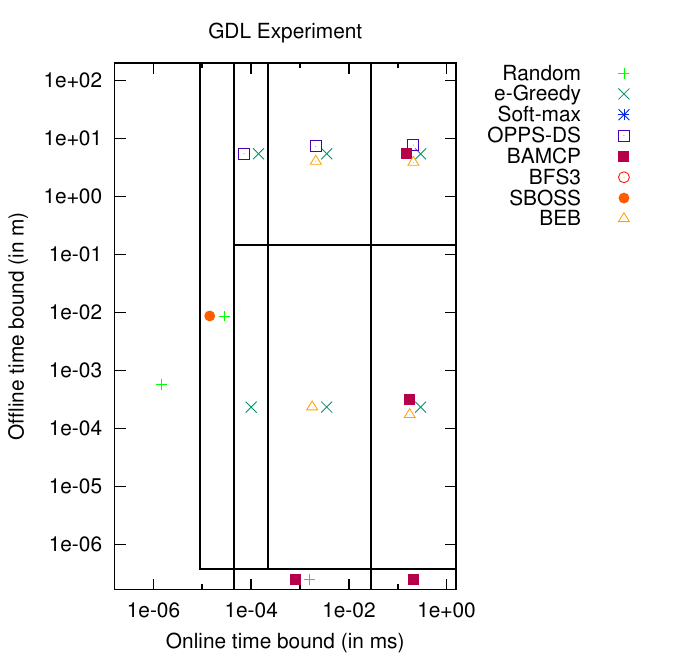}\\
\includegraphics[height=6cm]{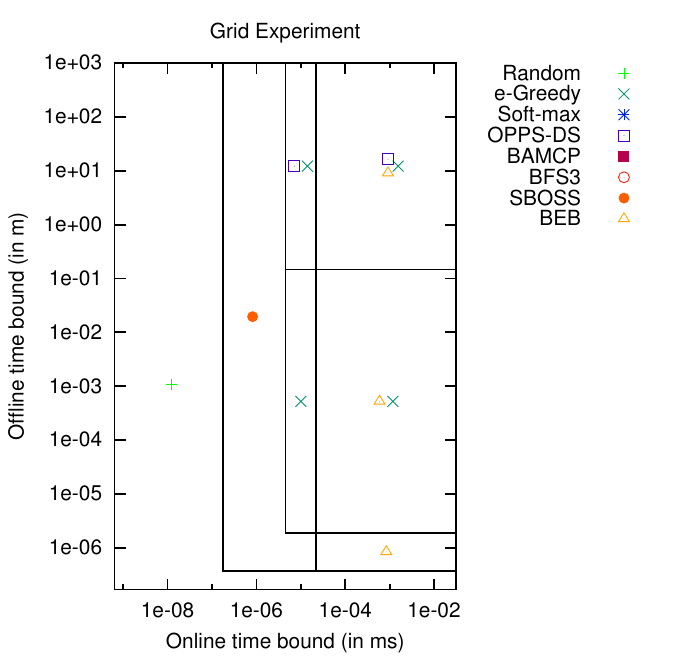}
\end{figure}
\caption{Best algorithms w.r.t offline/online time periods}
\label{exp_best_accurate}
\end{minipage}\hfill
\begin{minipage}{0.45\textwidth}
\centering
\begin{figure}[H]
\centering
\begin{small} \textbf{GC Experiment} \end{small}
{ \tiny
\begin{tabular}{p{5cm}HH|p{1.35cm}}
	\textbf{Agent} & \textbf{Offline time} & \textbf{Mean online time (per decision)} & \textbf{Score}\\
	\hline
	Random & $\boldsymbol{\sim 0}$\textbf{ms} & $\boldsymbol{\sim 0}$\textbf{ms} & $31.12 \pm 0.9$\\
	e-Greedy ($\epsilon$ = 0) & $\boldsymbol{\sim 0}$\textbf{ms} & $\boldsymbol{\sim 0}$\textbf{ms} & $40.62 \pm 1.55$\\
	Soft-Max ($\tau$ = 0.1) & $\boldsymbol{\sim 0}$\textbf{ms} & $\boldsymbol{\sim 0}$\textbf{ms} & $34.73 \pm 1.74$\\
	\textbf{OPPS-DS ($\boldsymbol{Q_2(x, u) / Q_0(x, u)}$)} & $\boldsymbol{\sim 50}$\textbf{s} & $\boldsymbol{\sim 0}$\textbf{ms} & $\boldsymbol{42.47 \pm 1.91}$\\
	BAMCP ($K$ = 2500, $depth$ = 15) & $\boldsymbol{\sim 0}$\textbf{ms} & $\boldsymbol{\sim 20}$\textbf{ms} & $35.56 \pm 1.27$\\
	BFS3 ($K$ = 500, $C$ = 15, $depth$ = 15) & $\boldsymbol{\sim 0}$\textbf{ms} & $\boldsymbol{\sim 45}$\textbf{ms} & $39.84 \pm 1.74$\\
	SBOSS ($\epsilon$ = 0.001, $\delta$ = 7) & $\boldsymbol{\sim 0}$\textbf{ms} & $\boldsymbol{\sim 0}$\textbf{ms} & $35.9 \pm 1.89$\\
	BEB ($\beta$ = 2.5) & $\boldsymbol{\sim 0}$\textbf{ms} & $\boldsymbol{\sim 0}$\textbf{ms} & $41.72 \pm 1.63$\\
\end{tabular}
}
\vspace{2cm}
\begin{small} \textbf{GDL Experiment} \end{small}
{ \tiny
\begin{tabular}{p{5cm}HH|p{1.35cm}}
	\textbf{Agent} & \textbf{Offline time} & \textbf{Mean online time (per decision)} & \textbf{Score}\\
	\hline
	Random & $\boldsymbol{\sim 0}$\textbf{ms} & $\boldsymbol{\sim 0}$\textbf{ms} & $2.79 \pm 0.07$\\
	e-Greedy ($\epsilon$ = 0.1) & $\boldsymbol{\sim 0}$\textbf{ms} & $\boldsymbol{\sim 0}$\textbf{ms} & $3.05 \pm 0.07$\\
	Soft-Max ($\tau$ = 0.1) & $\boldsymbol{\sim 0}$\textbf{ms} & $\boldsymbol{\sim 0}$\textbf{ms} & $2.79 \pm 0.1$\\
	OPPS-DS ($\max(Q_0(x, u), |Q_2(x,u)|)$) & $\boldsymbol{\sim 4}$\textbf{h} & $\boldsymbol{\sim 0}$\textbf{ms} & $3.1 \pm 0.07$\\
	\textbf{BAMCP ($\boldsymbol{K}$ = 10000, $\boldsymbol{depth}$ = 15)} & $\boldsymbol{\sim 0}$\textbf{ms} & $\boldsymbol{\sim 263}$\textbf{ms} & $\boldsymbol{3.11 \pm 0.07}$\\
	BFS3 ($K$ = 1, $C$ = 15, $depth$ = 25) & $\boldsymbol{\sim 0}$\textbf{ms} & $\boldsymbol{\sim 1}$\textbf{ms} & $2.9 \pm 0.07$\\
	SBOSS ($\epsilon$ = 1, $\delta$ = 1) & $\boldsymbol{\sim 0}$\textbf{ms} & $\boldsymbol{\sim 0}$\textbf{ms} & $2.81 \pm 0.1$\\
	BEB ($\beta$ = 0.5) & $\boldsymbol{\sim 0}$\textbf{ms} & $\boldsymbol{\sim 0}$\textbf{ms} & $3.09 \pm 0.07$\\
\end{tabular}
}

\vspace{2cm}
\begin{small} \textbf{Grid Experiment} \end{small}
{ \tiny
\begin{tabular}{p{5cm}HH|p{1.35cm}}
	\textbf{Agent} & \textbf{Offline time} & \textbf{Mean online time (per decision)} & \textbf{Score}\\
	\hline
	Random & $\boldsymbol{\sim 0}$\textbf{ms} & $\boldsymbol{\sim 0}$\textbf{ms} & $0.22 \pm 0.06$\\
	e-Greedy ($\epsilon$ = 0) & $\boldsymbol{\sim 0}$\textbf{ms} & $\boldsymbol{\sim 0}$\textbf{ms} & $6.9 \pm 0.31$\\
	Soft-Max ($\tau$ = 0.05) & $\boldsymbol{\sim 0}$\textbf{ms} & $\boldsymbol{\sim 0}$\textbf{ms} & $0 \pm 0$\\
	\textbf{OPPS-DS ($\boldsymbol{Q_0(x, u) + Q_2(x, u)}$)} & $\boldsymbol{\sim 13}$\textbf{m} & $\boldsymbol{\sim 1}$\textbf{ms} & $\boldsymbol{7.03 \pm 0.3}$\\
	BAMCP ($K$ = 25000, $depth$ = 15) & $\boldsymbol{\sim 0}$\textbf{ms} & $\boldsymbol{\sim 7}$\textbf{s} & $6.43 \pm 0.3$\\
	BFS3 ($K$ = 500, $C$ = 15, $depth$ = 50) & $\boldsymbol{\sim 0}$\textbf{ms} & $\boldsymbol{\sim 428}$\textbf{ms} & $3.46 \pm 0.23$\\
	SBOSS ($\epsilon$ = 0.1, $\delta$ = 7) & $\boldsymbol{\sim 0}$\textbf{ms} & $\boldsymbol{\sim 0}$\textbf{ms} & $4.5 \pm 0.33$\\
	BEB ($\beta$ = 0.5) & $\boldsymbol{\sim 0}$\textbf{ms} & $\boldsymbol{\sim 1}$\textbf{ms} & $6.76 \pm 0.3$\\
\end{tabular}
}

\vspace{1.7cm}
\end{figure}
\caption{Best algorithms w.r.t Performance}
\label{exp_best_performance_accurate}
\end{minipage}
\end{figure}

\newpage
As it can be seen in Figure~\ref{exp_offline_accurate}, OPPS is the only algorithm whose offline time cost varies. 
In the three different settings, OPPS can be launched after a few seconds, but behaves very poorly. 
However, its performances increased very quickly when given at least one minute of computation time.
Algorithms that do not use offline computation time have a wide range of different scores.
This variance represents the different possible configurations for these algorithms, which only lead to different online computation time.

On Figure~\ref{exp_online_accurate}, BAMCP, BFS3 and SBOSS have variable online time costs. 
BAMCP behaved poorly on the first experiment, but obtained the best score on the second one and was pretty efficient on the last one. 
BFS3 was good only on the second experiment. 
SBOSS was never able to get a good score in any cases. 
Note that OPPS online time cost varies slightly depending on the formula's complexity.

If we take a look at the top-right point in Figure~\ref{exp_best_accurate}, which defines the less restrictive bounds, we notice that OPPS-DS and BEB were always the best algorithms in every experiment.
$\epsilon$-Greedy was a good candidate in the two first experiments.
BAMCP was also a very good choice except for the first experiment.
On the contrary, BFS3 and SBOSS were only good choices in the first experiment.

If we look closely, we can notice that OPPS-DS was always one of the best algorithm since we have met its minimal offline computation time requirements.

Moreover, when we place our offline-time bound right under OPPS-DS minimal offline time cost, we can see how the top is affected from left to right:
\begin{itemize}
	\item[]
	\begin{tabular}{p{1cm}l}
		\textbf{GC}: &(Random), (SBOSS), (BEB, $\epsilon$-Greedy), (BEB, BFS3, $\epsilon$-Greedy),
	\end{tabular}

	\item[]
	\begin{tabular}{p{1cm}l}
		\textbf{GDL}: &(Random), (Random, SBOSS), ($\epsilon$-Greedy), (BEB, $\epsilon$-Greedy), \\
		&(BAMCP, BEB, $\epsilon$-Greedy),
	\end{tabular}
	
	\item[]
	\begin{tabular}{p{1cm}l}
		\textbf{Grid}: &(Random), (SBOSS), ($\epsilon$-Greedy), (BEB, $\epsilon$-Greedy).
	\end{tabular}
\end{itemize}
We can clearly see that SBOSS was the first algorithm to appear on the top, with a very small online computation cost, followed by $\epsilon$-Greedy and BEB.
Beyond a certain online time bound, BFS3 emerged in the first experiment while BAMCP emerged in the second experiment.
Neither of them was able to compete with BEB or $\epsilon$-Greedy in the last experiment.

Soft-max was never able to reach the top regardless the configuration.

Figure~\ref{exp_best_performance_accurate} reports the best score observed for each algorithm, disassociated from any time measure. Note that the variance is very similar for all algorithms in GDL and Grid experiments. On the contrary, the variance oscillates between $1.0$ and $2.0$. However, OPPS seems to be the less stable algorithm in the three cases.

\newpage
\subsubsection{Inaccurate case}
\begin{figure}[H]
\centering
\begin{minipage}{0.45\textwidth}
\centering
\begin{figure}[H]
\centering
\includegraphics[width=\textwidth]{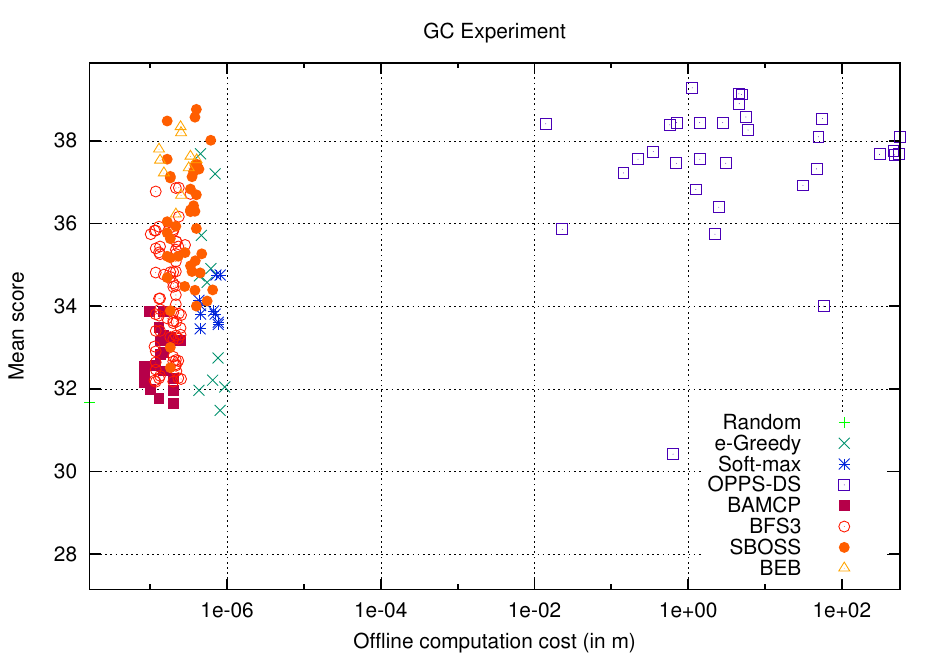}\\
\includegraphics[width=\textwidth]{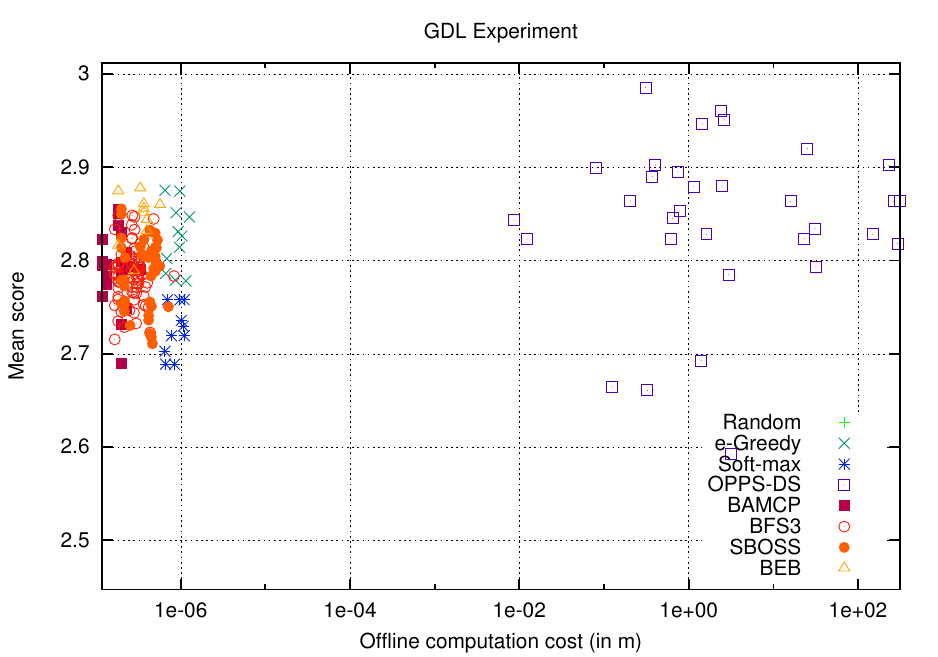}\\
\includegraphics[width=\textwidth]{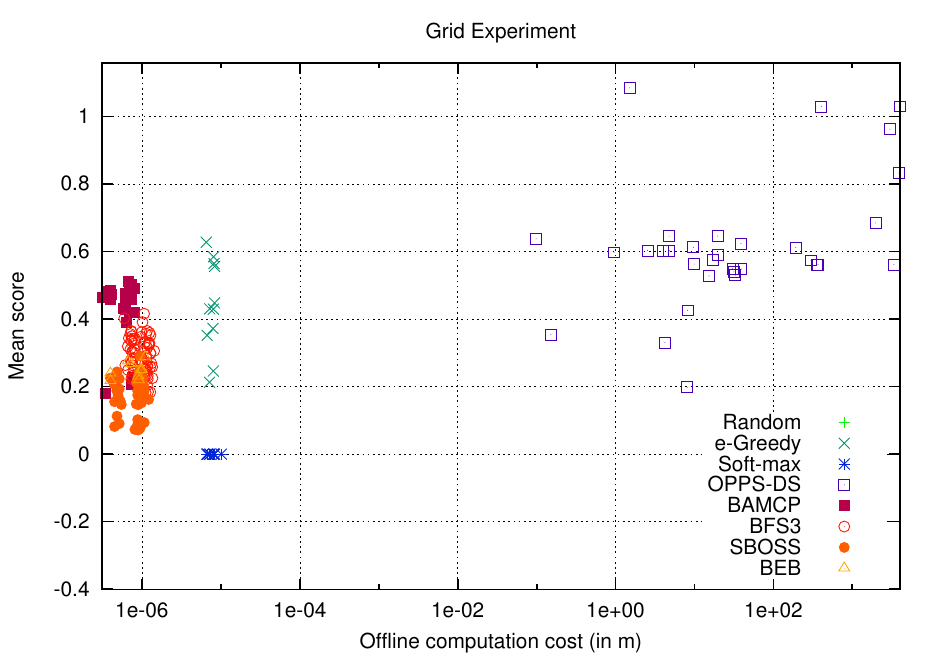}
\end{figure}
\caption{Offline computation cost Vs. Performance}
\label{exp_offline_inaccurate}
\end{minipage}\hfill
\begin{minipage}{0.45\textwidth}
\centering
\begin{figure}[H]
\centering
\includegraphics[width=\textwidth]{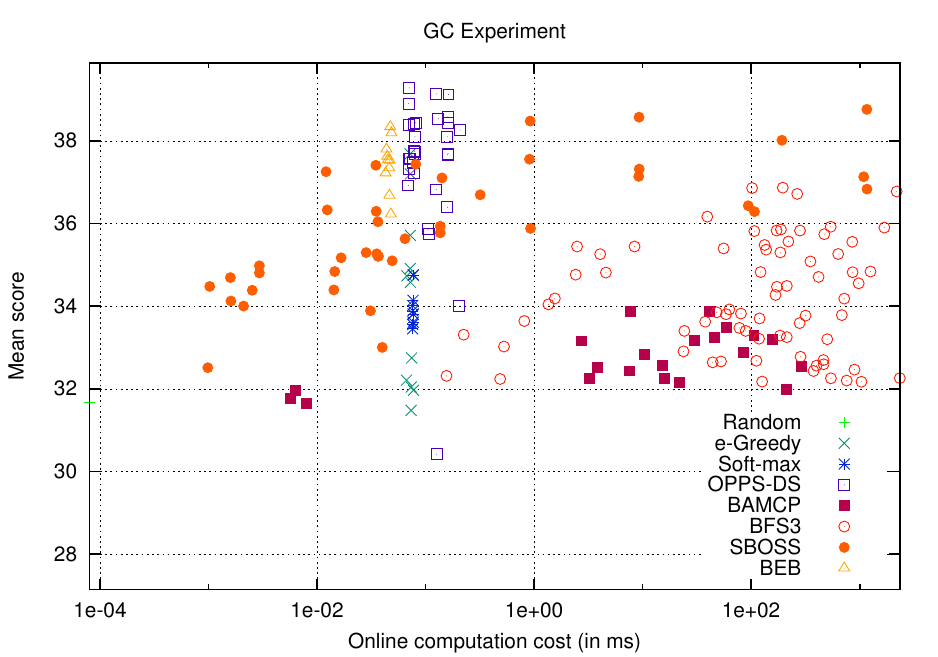}\\
\includegraphics[width=\textwidth]{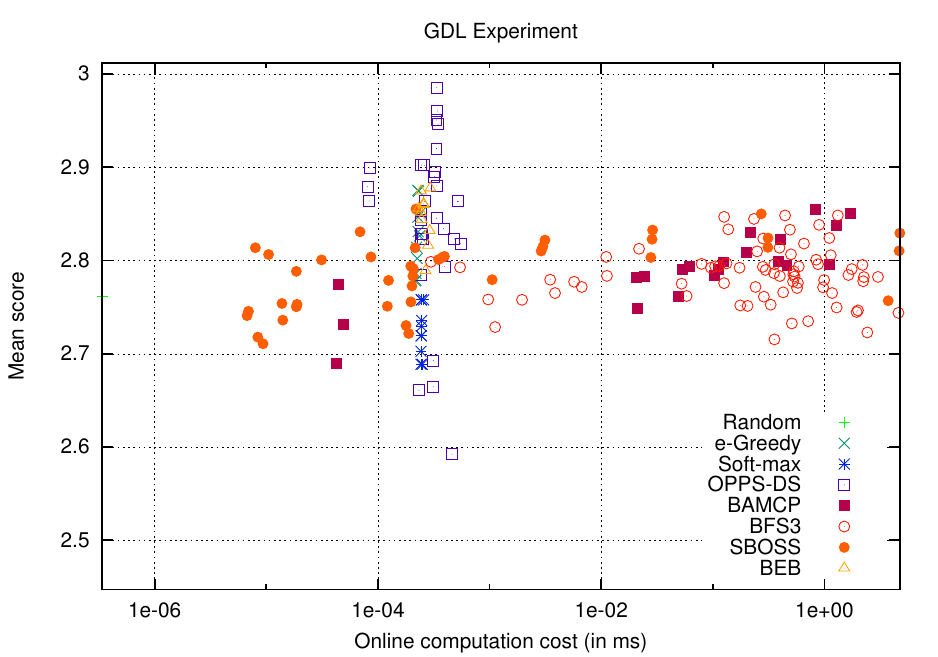}\\
\includegraphics[width=\textwidth]{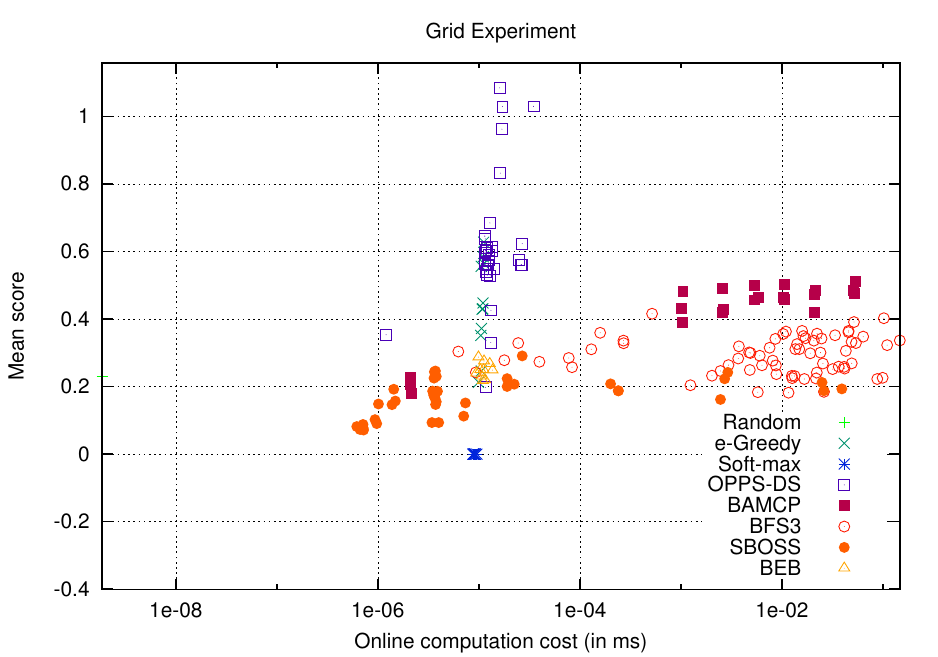}
\end{figure}
\caption{Online computation cost Vs. Performance}
\label{exp_online_inaccurate}
\end{minipage}
\end{figure}

\newpage
\begin{figure}[H]
\centering
\begin{minipage}{0.45\textwidth}
\centering
\begin{figure}[H]
\centering
\includegraphics[height=6cm]{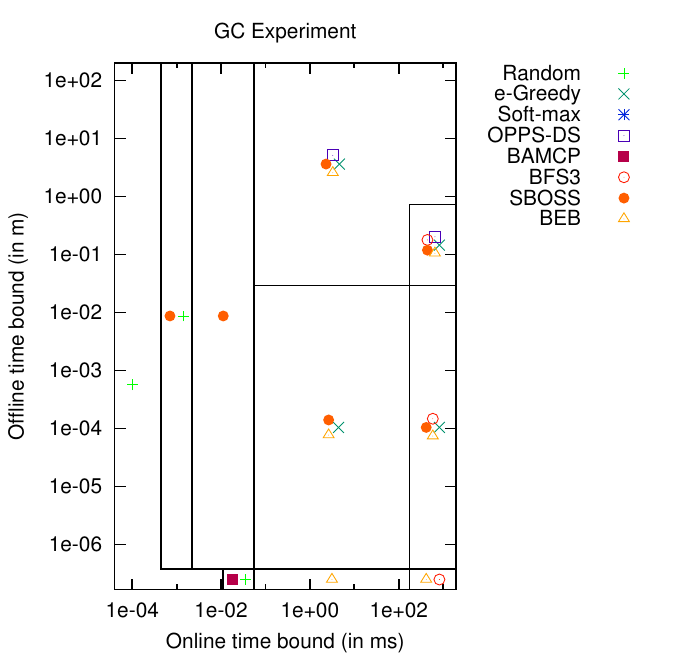}\\
\includegraphics[height=6cm]{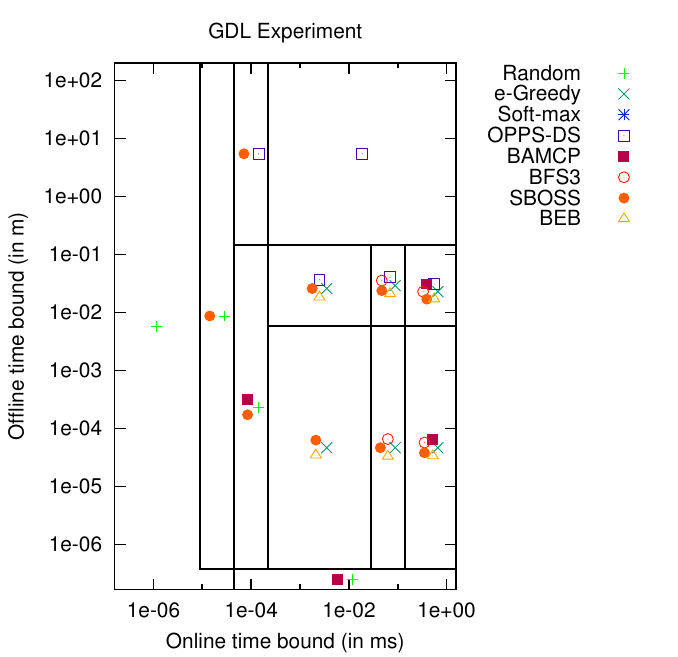}\\
\includegraphics[height=6cm]{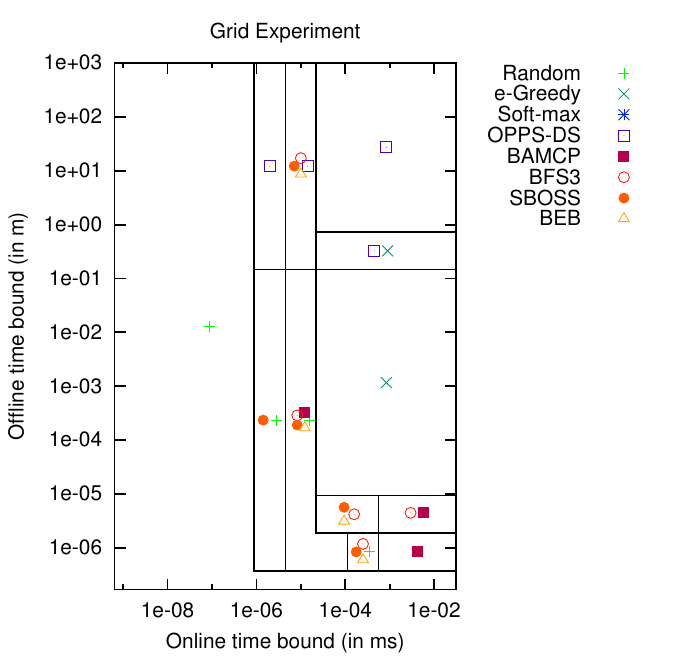}
\end{figure}
\caption{Best algorithms w.r.t offline/online time periods}
\label{exp_best_inaccurate}
\end{minipage}\hfill
\begin{minipage}{0.45\textwidth}
\centering
\begin{figure}[H]
\centering
\begin{small} \textbf{GC Experiment}\\ \end{small}
{ \tiny
\begin{tabular}{p{5cm}HH|p{1.35cm}}
	\textbf{Agent} & \textbf{Offline time} & \textbf{Mean online time (per decision)} & \textbf{Score}\\
	\hline
	Random & $\boldsymbol{\sim 0}$\textbf{ms} & $\boldsymbol{\sim 0}$\textbf{ms} & $31.67 \pm 1.05$\\
	e-Greedy ($\epsilon$ = 0) & $\boldsymbol{\sim 0}$\textbf{ms} & $\boldsymbol{\sim 0}$\textbf{ms} & $37.69 \pm 1.75$\\
	Soft-Max ($\tau$ = 0.33) & $\boldsymbol{\sim 0}$\textbf{ms} & $\boldsymbol{\sim 0}$\textbf{ms} & $34.75 \pm 1.64$\\
	\textbf{OPPS-DS ($\boldsymbol{Q_{0}(x, u)}$)} & $\boldsymbol{\sim 1}$\textbf{m} & $\boldsymbol{\sim 0}$\textbf{ms} & $\boldsymbol{39.29 \pm 1.71}$\\
	BAMCP ($K$ = 1250, $depth$ = 25) & $\boldsymbol{\sim 0}$\textbf{ms} & $\boldsymbol{\sim 8}$\textbf{ms} & $33.87 \pm 1.26$\\
	BFS3 ($K$ = 1250, $C$ = 15, $depth$ = 25) & $\boldsymbol{\sim 0}$\textbf{ms} & $\boldsymbol{\sim 196}$\textbf{ms} & $36.87 \pm 1.82$\\
	SBOSS ($\epsilon$ = 1e-06, $\delta$ = 0.0001) & $\boldsymbol{\sim 0}$\textbf{ms} & $\boldsymbol{\sim 1}$\textbf{s} & $38.77 \pm 1.89$\\
	BEB ($\beta$ = 16) & $\boldsymbol{\sim 0}$\textbf{ms} & $\boldsymbol{\sim 0}$\textbf{ms} & $38.34 \pm 1.62$\\
\end{tabular}
}
\vspace{2cm}
\begin{small} \textbf{GDL Experiment} \end{small}
{ \tiny
\begin{tabular}{p{5cm}HH|p{1.35cm}}
	\textbf{Agent} & \textbf{Offline time} & \textbf{Mean online time (per decision)} & \textbf{Score}\\
	\hline
	Random & $\boldsymbol{\sim 0}$\textbf{ms} & $\boldsymbol{\sim 0}$\textbf{ms} & $2.76 \pm 0.08$\\
	e-Greedy ($\epsilon$ = 0.3) & $\boldsymbol{\sim 0}$\textbf{ms} & $\boldsymbol{\sim 0}$\textbf{ms} & $2.88 \pm 0.07$\\
	Soft-Max ($\tau$ = 0.05) & $\boldsymbol{\sim 0}$\textbf{ms} & $\boldsymbol{\sim 0}$\textbf{ms} & $2.76 \pm 0.1$\\
	\textbf{OPPS-DS ($\boldsymbol{\max(Q_0(x, u), Q_1(x, u)}$)} & $\boldsymbol{\sim 19}$\textbf{s} & $\boldsymbol{\sim 0}$\textbf{ms} & $\boldsymbol{2.99 \pm 0.08}$\\
	BAMCP ($K $= 10000, $depth$ = 50) & $\boldsymbol{\sim 0}$\textbf{ms} & $\boldsymbol{\sim 207}$\textbf{ms} & $2.85 \pm 0.07$\\
	BFS3 ($K$ = 1250, C = 15, $depth$ = 50) & $\boldsymbol{\sim 0}$\textbf{ms} & $\boldsymbol{\sim 329}$\textbf{ms} & $2.85 \pm 0.07$\\
	SBOSS ($\epsilon$ = 0.1, $\delta$ = 0.001) & $\boldsymbol{\sim 0}$\textbf{ms} & $\boldsymbol{\sim 0}$\textbf{ms} & $2.86 \pm 0.07$\\
	BEB ($\beta$ = 2.5) & $\boldsymbol{\sim 0}$\textbf{ms} & $\boldsymbol{\sim 0}$\textbf{ms} & $2.88 \pm 0.07$\\
\end{tabular}
}

\vspace{2cm}
\begin{small} \textbf{Grid Experiment} \end{small}
{ \tiny
\begin{tabular}{p{5cm}HH|p{1.35cm}}
	\textbf{Agent} & \textbf{Offline time} & \textbf{Mean online time (per decision)} & \textbf{Score}\\
	\hline
	Random & $\boldsymbol{\sim 0}$\textbf{ms} & $\boldsymbol{\sim 0}$\textbf{ms} & $0.23 \pm 0.06$\\
	e-Greedy ($\epsilon$ = 0.2) & $\boldsymbol{\sim 0}$\textbf{ms} & $\boldsymbol{\sim 1}$\textbf{ms} & $0.63 \pm 0.09$\\
	Soft-Max ($\tau$ = 0.05) & $\boldsymbol{\sim 0}$\textbf{ms} & $\boldsymbol{\sim 1}$\textbf{ms} & $0 \pm 0$\\
	\textbf{OPPS-DS ($\boldsymbol{Q_1(x, u) + Q_2(x, u)}$))} & $\boldsymbol{\sim 2}$\textbf{m} & $\boldsymbol{\sim 1}$\textbf{ms} & $\boldsymbol{1.09 \pm 0.17}$\\
	BAMCP ($K$ = 25000, $depth$ = 25) & $\boldsymbol{\sim 0}$\textbf{ms} & $\boldsymbol{\sim 3}$\textbf{s} & $0.51 \pm 0.09$\\
	BFS3 ($K$ = 1, $C$ = 15, $depth$ = 50) & $\boldsymbol{\sim 0}$\textbf{ms} & $\boldsymbol{\sim 32}$\textbf{ms} & $0.42 \pm 0.09$\\
	SBOSS ($\epsilon$ = 0.001, $\delta$ = 0.1) & $\boldsymbol{\sim 0}$\textbf{ms} & $\boldsymbol{\sim 2}$\textbf{ms} & $0.29 \pm 0.07$\\
	BEB ($\beta$ = 0.25) & $\boldsymbol{\sim 0}$\textbf{ms} & $\boldsymbol{\sim 1}$\textbf{ms} & $0.29 \pm 0.05$\\
\end{tabular}
}

\vspace{1.7cm}
\end{figure}
\caption{Best algorithms w.r.t Performance}
\label{exp_best_performance_inaccurate}
\end{minipage}
\end{figure}

\newpage
As seen in the accurate case, Figure~\ref{exp_offline_inaccurate} also shows impressive performances for OPPS-DS, which has beaten all other algorithms in every experiment.
We can also notice that, as observed in the accurate case, in the Grid experiment, the OPPS-DS agents scores are very close.
However, only a few were able to significantly surpass the others, contrary to the accurate case where most OPPS-DS agents were very good candidates.

Surprisingly, SBOSS was a very good alternative to BAMCP and BFS3 in the two first experiments as shown in Figure~\ref{exp_online_inaccurate}.
It was able to surpass both algorithms on the first one while being very close to BAMCP performances in the second.
Relative performances of BAMCP and BFS3 remained the same in the inaccurate case, even if the BAMCP advantage is less visible in the second experiment.
BEB was no longer able to compete with OPPS-DS and was even beaten by BAMCP and BFS3 in the last experiment.
$\epsilon$-Greedy was still a decent choice except in the first experiment.
As observed in the accurate case, Soft-max was very bad in every case.

In Figure~\ref{exp_best_inaccurate}, if we take a look at the top-right point, we can see OPPS-DS is the best choice in the second and third experiment.
BEB, SBOSS and $\epsilon$-Greedy share the first place with OPPS-DS in the first one.

If we place our offline-time bound right under OPPS-DS minimal offline time cost, we can see how the top is affected from left to right:
\begin{itemize}
	\item[]
	\begin{tabular}{p{1cm}l}
		\textbf{GC}: &(Random), (Random, SBOSS), (SBOSS), (BEB, SBOSS, $\epsilon$-Greedy),\\
		&(BEB, BFS3, SBOSS, $\epsilon$-Greedy),
	\end{tabular}

	\item[]
	\begin{tabular}{p{1cm}l}
		\textbf{GDL}: &(Random), (Random, SBOSS), (BAMCP, Random, SBOSS), \\
		&(BEB, SBOSS, $\epsilon$-Greedy), (BEB, BFS3, SBOSS, $\epsilon$-Greedy), \\
		&(BAMCP, BEB, BFS3, SBOSS, $\epsilon$-Greedy),
	\end{tabular}

	\item[]
	\begin{tabular}{p{1cm}l}
		\textbf{Grid}: &(Random), (Random, SBOSS), (BAMCP, BEB, BFS3, Random, SBOSS),\\
		&($\epsilon$-Greedy).
	\end{tabular}
\end{itemize}

SBOSS is again the first algorithm to appear in the rankings.
$\epsilon$-Greedy is the only one which could reach the top in every case, even when facing BAMCP and BFS3 fed with high online computation cost.
BEB no longer appears to be undeniably better than the others.
Besides, the two first experiments show that most algorithms obtained similar results, except for BAMCP which does not appear on the top in the first experiment.
In the last experiment, $\epsilon$-Greedy succeeded to beat all other algorithms.

Figure~\ref{exp_best_performance_inaccurate} does not bring us more information than those we observed in the accurate case.

\subsubsection{Summary}
In the accurate case, OPPS-DS was always among the best algorithms, at the cost of some offline computation time.
When the offline time budget was too constrained for OPPS-DS, different algorithms were suitable depending on the online time budget:
\begin{itemize}
	\item	\textbf{Low online time budget:} SBOSS was the fastest algorithm to make better decisions than a random policy.
	\item	\textbf{Medium online time budget\footnote{$\pm$ 100 times more than the low online time budget}:} BEB reached performances similar to OPPS-DS on each experiment.
	\item	\textbf{High online time budget\footnote{$\pm$ 100 times more than the medium online time budget}:} In the first experiment, BFS3 managed to catch up BEB and OPPS-DS when given sufficient time.
	In the second experiment, it was BAMCP which has achieved this result.
	Neither BFS3 nor BAMCP was able to compete with BEB and OPPS-DS in the last experiment.
\end{itemize}

The results obtained in the inaccurate case were very interesting.
BEB was not as good as it seemed to be in the accurate case, while SBOSS improved significantly compared to the others.
For its part, OPPS-DS obtained the best overall results in the inaccurate case by outperforming all the other algorithms in two out of three experiments while remaining among the best ones in the last experiment.

\section{Conclusion}
\label{conclusion}
We have proposed a new extensive BRL comparison methodology which takes into account both performance and time requirements for each algorithm.
In particular, our benchmarking protocol shows that no single algorithm dominates all other algorithms on all scenarios.
The protocol we introduced can compare any time algorithm to non-anytime algorithms while measuring the impact of inaccurate offline training.
By comparing algorithms on large sets of problems, we avoid over fitting to a single problem.
Our methodology is associated with an open-source library, BBRL, and we hope that it will help other researchers to design algorithms whose performances are put into perspective with computation times, that may be critical in many applications.
This library is specifically designed to handle new algorithms easily, and is provided with a complete and comprehensive documentation website.

\acks{Micha\"el Castronovo acknowledges the financial support of the FRIA. Raphael Fonteneau is a postdoctoral fellow of the F.R.S.-FNRS (Belgian Funds for Scientifique Research).}

\newpage
\appendix
\section{Pseudo-code of the algorithms}
\label{appendix:pseudocode}
\hspace{0.4cm}
\begin{minipage}[t]{.9\textwidth}
\begin{algorithm}[H]
\caption{$\epsilon$-Greedy}\label{eGreedyPseudocode}
\begin{algorithmic}[1]
\Procedure{offline-learning}{$p^{0}_{\mathcal M}(.)$}
	\State $\hat{M} \leftarrow $ ``Build an initial model based on $p^{0}_{\mathcal M}(.)$''
\EndProcedure
\State
\Function{search}{$x, h$}
	\State \{Draw a random value in $[0; 1]$\}
	\State $r \leftarrow \mathcal{U}(0, 1)$
	\State
	\If{$r < \epsilon$} \; \{Random case\}
		\State \textbf{return} ``An action selected randomly''
	\State
	\Else \; \{Greedy case\}
		\State $\pi^{*}_{\hat{M}} \leftarrow $\Call{value-iteration}{$\hat{M}$}
		\State \textbf{return} $\pi^{*}_{\hat{M}}(x)$
	\EndIf
\EndFunction
\State
\Procedure{online-learning}{$x, u, y, r$}
	\State ``Update model $\hat{M}$ w.r.t. transition $<x, u, y, r>$''
\EndProcedure
\end{algorithmic}
\end{algorithm}
\vspace{9.5cm}
\end{minipage}

\begin{minipage}[t]{.9\textwidth}
\begin{algorithm}[H]
\caption{Soft-max}\label{SoftMaxPseudocode}
\begin{algorithmic}[1]
\Procedure{offline-learning}{$p^{0}_{\mathcal M}(.)$}
	\State $\hat{M} \leftarrow $ ``Build an initial model based on $p^{0}_{\mathcal M}(.)$''
\EndProcedure
\State
\Function{search}{$x, h$}
	\State \{Draw a random value in $[0; 1]$\}
	\State $r \leftarrow \mathcal{U}(0, 1)$
	\State
	\State \{Select an action randomly, with a probability proportional to $Q_{\hat{M}}^{*}(x, u)$\}
	\State	$Q_{\hat{M}}^{*} \leftarrow $ ``Compute the optimal Q-function of $\hat{M}$''
	\For{$1 \leq i \leq |U|$}
		\If{$r < \sum_{j \leq i} \frac{\exp\left( Q_{\hat{M}}^{*}(x, u^{(j)}) \middle/ \tau \right)}{\sum_{u'} \exp\left( Q_{\hat{M}}^{*}(x, u') \middle/ \tau \right)}$}
			\State \textbf{return} $u^{(i)}$
		\EndIf
	\EndFor
\EndFunction
\State
\Procedure{online-learning}{$x, u, y, r$}
	\State ``Update model $\hat{M}$ w.r.t. transition $<x, u, y, r>$''
\EndProcedure
\end{algorithmic}
\end{algorithm}
\end{minipage}

\begin{minipage}[t]{.9\textwidth}
\begin{algorithm}[H]
\caption{OPPS-DS}\label{OPPSpseudocode}
\begin{algorithmic}[1]
\Procedure{offline-learning}{$p^{0}_{\mathcal M}(.)$}
  	\State \{Initialise the $k$ arms of UCB1\}
  	\For{$1 \leq i \leq k$}
		\State $M \sim p^{0}_{\mathcal M}(.)$
  		\State $R^{\pi_i}_M \leftarrow$ ``Simulate strategy $\pi_i$ on MDP $M$ over a single trajectory''
  		\State $\mu(i) \leftarrow R^{\pi_i}_M$
  		\State $\theta(i) \leftarrow 1$
	\EndFor
	\State
	\State \{Run UCB1 with a budget of $\beta$\}
	\For{$k+1 \leq b \leq \beta$}
 		\State $a \leftarrow \argmax_{a'} \mu(a') + \sqrt{\frac{2 \log(b)}{\theta(a')}}$
 		\State $M \sim p^{0}_{\mathcal M}(.)$
 		\State $R^{\pi_a}_M \leftarrow$ ``Simulate strategy $\pi_a$ on MDP $M$ over a single trajectory''
 		\State $\mu(a) \leftarrow \frac{\theta(a) \mu(a) + R^{\pi_a}_M}{\theta(a) + 1}$
 		\State $\theta(a) \leftarrow \theta(a) + 1$
 	\EndFor
  	\State
  	\State \{Select the E/E strategy associated to the most drawn arm\}
  	\State $a^{*} \leftarrow \argmax_{a'} \theta(a')$
  	\State $\pi_{OPPS} \leftarrow \pi_{a^{*}}$
\EndProcedure
\State
\Function{search}{$x, h$}
	\State \textbf{return} $u \sim \pi_{OPPS}(x, h)$
\EndFunction
\State
\Procedure{online-learning}{$x, u, y, r$}
	\State ``Update strategy $\pi_{OPPS}$ w.r.t. transition $<x, u, y, r>$''
\EndProcedure
\end{algorithmic}
\end{algorithm}
\end{minipage}

\begin{minipage}[t]{.9\textwidth}
\begin{algorithm}[H]
\caption{BAMCP (1/2)}\label{BAMCPpseudocode}
\begin{algorithmic}[1]
\Function{search}{$x, h$}
	\State \{Develop a MCTS and compute $Q(., .)$\}
	\For{$1 \leq k \leq K$}
		\State $M \sim p^{h}_{\mathcal M}$
		\State \Call{Simulate}{$\langle x,h\rangle,M,0$}
	\EndFor
	\State
	\State \{Return the best action w.r.t. $Q(., .)$\}
	\State \textbf{return} $\argmax_u Q(\langle x,h\rangle,u)$
\EndFunction
\State
\Function{simulate}{$\langle x,h\rangle,M,d$}
\If{$N(\langle x,h\rangle)=0$} \{New node reached\}
	\State ``Initialise $N(\langle x,h\rangle, u)$, $Q(\langle x,h\rangle, u)$''
	\State $u \sim \pi_{0}(\langle x,h\rangle)$
	\State ``Sample $x',r$ from model $M$''
	\State
	\State \{Estimate the score of this node by using the rollout policy\}
	\State $R \leftarrow r + \gamma$ \Call{Rollout}{$\langle x',hux'\rangle,P,d$}
	\State
	\State ``Update $N(\langle x,h\rangle)$, $N(\langle x,h\rangle,u)$, $Q(\langle x,h\rangle,u)$''
	\State \textbf{return} $R$
\EndIf
	\State
    \State \{Select the next branch to explore\}
	\State $u \leftarrow \argmax_{u'} Q(\langle x,h\rangle,u) + c \sqrt(\frac{log(N(\langle x,h\rangle))}{N(\langle x,h\rangle,u')})$
	\State ``Sample $x',r$ from model $M$''
	\State
	\State \{Follow the branch and evaluate it\}
	\State $R \leftarrow r + \gamma$ \Call{Simulate}{$\langle x',hux'\rangle,M,d+1$}
	\State 
	\State ``Update $N(\langle x,h\rangle)$, $N(\langle x,h\rangle,u)$, $Q(\langle x,h\rangle,u)$''
	\State \textbf{return} $R$
\EndFunction
\end{algorithmic}
\end{algorithm}
\end{minipage}

\begin{minipage}[t]{.9\textwidth}
\begin{algorithm}[H]
\caption{BAMCP (2/2)}\label{BAMCPpseudocode}
\begin{algorithmic}[1]
\Procedure{rollout}{$\langle x,h\rangle,M,d$}
	\If{$\gamma^{d} R_{max} < \epsilon$} \{Truncate the trajectory if precision $\epsilon$ has been reached\}
	\State \textbf{return} $0$
	\EndIf
	\State
	\State \{Use the rollout policy to choose the action to perform\}
	\State $u \sim \pi_{0}(x,h)$
	\State
	\State \{Simulate a single transition from $M$ and continue the rollout process\}
	\State $y \sim P_{M}$
	\State $r \leftarrow \rho_M(x, u, y)$
	\State \textbf{return} $r + \gamma \; $\Call{rollout}{$\langle y, huy\rangle, M, d + 1$}
\EndProcedure
\State
\Procedure{online-learning}{$x, u, y, r$}
	\State ``Update the posterior w.r.t. transition $<x, u, y, r>$''
\EndProcedure
\end{algorithmic}
\end{algorithm}
\end{minipage}

\newpage
\begin{minipage}[t]{.9\textwidth}
\begin{algorithm}[H]
\caption{BFS3}\label{BFS3pseudocode}
\begin{algorithmic}[1]
\Function{search}{$x, h$}
\State \{Update the current Q-function\}
\State $M_{mean} \leftarrow $ ``Compute the mean MDP of $p^{t}_{\mathcal M}(.)$.''
\ForAll{$u \in$ U}
	\For{$1 \leq i \leq C$}
		\State \{Draw $y$ and $r$ from the mean MDP of the posterior\}
		\State $y \sim P_{M_{mean}}$
		\State $r \leftarrow \rho_M(x, u, y)$
		\State
		\State \{Update the Q-value in $(x, u)$ by using $FSSS$ algorithm\}
		\State $Q(x, u) \leftarrow Q(x, u) + \frac{1}{C} \big[ r+\gamma$ \Call{FSSS}{$y, d, t$} $\big]$
	\EndFor
\EndFor
\State
\State \{Return the action $u$ with the maximal Q-value in $x$\}
\State \textbf{return} $\argmax_u Q(x, u)$
\EndFunction
\end{algorithmic}
\end{algorithm}
\end{minipage}

\begin{minipage}[t]{.9\textwidth}
\begin{algorithm}[H]
\caption{FSSS (1/2)}\label{FSSSpseudocode}
\begin{algorithmic}[1]
\Function{FSSS}{$x, d, t$}
\State \{Develop a MCTS and compute bounds on $V(x)$\}
\For{$1 \leq i \leq t$}
	\State \Call{rollout}{$s, d, 0$}
\EndFor
\State
\State \{Make an optimistic estimation of $V(x)$\}
\State $\hat{V}(x) \leftarrow \max_u U_d(x,u)$
\State \textbf{return} $\hat{V}(x)$
\EndFunction
\end{algorithmic}
\end{algorithm}
\vspace{5cm}
\end{minipage}

\begin{minipage}[t]{.9\textwidth}
\begin{algorithm}[H]
\caption{FSSS (2/2)}\label{FSSSpseudocode}
\begin{algorithmic}[1]
\Procedure{rollout}{x, d, l}
\If{$d = l$} \{Stop when reaching the maximal depth\}
	\State \textbf{return}
\EndIf
\State
\If{$\neg Visited_d(x)$} \{New node reached\}
	\State \{Initialise this node\}
	\ForAll{$u \in U$}
		\State ``Initialise $N_d(x,u,x')$,$R_d(x,u)$''
		\For{$1 \leq i \leq C$}
			\State ``Sample $x',r$ from M''
			\State ``Update $N_d(x,u,x')$,$R_d(x,u)$''
			\State
			\If{$\neg Visited_d(x')$}
				\State $U_{d+1}(x'), L_{d+1}(x') = V_{max}, V_{min}$
			\EndIf
		\EndFor
	\EndFor
	\State
	\State \{Back-propagate this node's information\}
	\State \Call{Bellman-backup}{$x,d$}
	\State
	\State $Visited_d(x) \leftarrow $true
\EndIf
\State
\State \{Select an action and simulate a transition optimistically\}
\State $u \leftarrow \argmax_u U_d(x,u)$
\State $x' \leftarrow \argmax_{x'} \big( U_{d+1}(x') - L_{d+1}(x')\big) N_d(x,u,x')$
\State
\State \{Continue the rollout process and back-propagate the result\}
\State \Call{rollout}{$x',d,l+1$}
\State \Call{Bellman-backup}{$x,d$}
\State \textbf{return}
\EndProcedure
\end{algorithmic}
\end{algorithm}
\end{minipage}

\begin{minipage}[t]{.9\textwidth}
\begin{algorithm}[H]
\caption{SBOSS (1/2)}\label{SBOSSpseudocode}
\begin{algorithmic}[1]
\Function{search}{$x, h$}
	\State \{Compute the transition matrix of the mean MDP of the posterior\}
	\State $M_{mean} \leftarrow $ ``Compute the mean MDP of $p^{t}_{\mathcal M}(.)$.''
	\State $P_t \leftarrow P_{M_{mean}}$
	\State
	\State \{Update the policy to follow if necessary\}
	\State $\forall (x, u): \Delta(x, u) = \sum_{y \in X} \frac{|P_t(x, u, y) - P_{lastUpdate}(x, u, y)|}{\sigma(x, u, y)}$
	\If{$t = 1$ or $\exists (x', u') : \Delta(x', u') > \delta$}
		\State \{Sample some transition vectors for each state-action pair\}
		\State $S \leftarrow \{\}$
		\ForAll{$(x, u) \in X \times U$}
			\State \{Compute the number of transition vectors to sample for $(x, u)$\}
			\State $K_t(x, u) \leftarrow \max_y \left\lceil \frac{\sigma^{2}(x, u, y)}{\epsilon} \right\rceil$
			\State
			\State \{Sample $K_t(x, u)$ transition vectors from $<x, u>$, sampled from the posterior\}
			\For{$1 \leq k \leq K_t(x, u)$}
				\State $S \leftarrow S \; \cup $ ``A transition vector from $<x, u>$, sampled from the posterior''
			\EndFor
		\EndFor
		\State
		\State $M^{\#} \leftarrow $ ``Build a new MDP by merging all transitions from $S$''
		\State $\pi^{*}_{M^{\#}} \leftarrow $\Call{value-iteration}{$M^{\#}$}
		\State $\pi_{SBOSS} \leftarrow $\Call{fit-action-space}{$\pi^{*}_{M^{\#}}$}
		\State $P_{lastUpdate} \leftarrow P_t$
	\EndIf
	\State
	\State \{Return the optimal action in $x$ w.r.t. $\pi_{SBOSS}$\}
	\State \textbf{return} $u \sim \pi_{SBOSS}(x)$
\EndFunction
\end{algorithmic}
\end{algorithm}
\end{minipage}

\begin{minipage}[t]{.9\textwidth}
\begin{algorithm}[H]
\caption{SBOSS (2/2)}\label{SBOSSpseudocode}
\begin{algorithmic}[1]
\Function{fit-action-space}{$\pi^{*}_{M^{\#}}$}
	\ForAll{$x \in X$}
		\State $\pi(x) \leftarrow \pi^{*}_{M^{\#}}(x) \mod |U|$
	\EndFor
	\State
	\State \textbf{return} $\; \pi$
\EndFunction
\State
\Procedure{online-learning}{$x, u, y, r$}
	\State ``Update the posterior w.r.t. transition $<x, u, y, r>$''
\EndProcedure
\end{algorithmic}
\end{algorithm}
\end{minipage}

\begin{minipage}[t]{.9\textwidth}
\begin{algorithm}[H]
\begin{algorithmic}[1]
  \Procedure{search}{$x, h$}
   \State $M \leftarrow $ ``Compute the mean MDP of $p^{t}_{\mathcal M}(.)$.''
   \State
   \State \{Add a bonus reward to all transitions\}
   \For{$<x, u, y> \in \mathcal{X} \times \mathcal{U} \times \mathcal{X}$}
   		$\rho_M(x, u, y)  \leftarrow \rho_M(x, u, y) + \frac{\beta}{c^{(t)}_{<x, u, y>}}$
   \EndFor
   \State
   \State \{Compute the optimal policy of the modified MDP\}
   \State $\pi^{*}_M \leftarrow $\Call{value-iteration}{$M$}
   \State
   \State \{Return the optimal action in $x$ w.r.t. $\pi^{*}_M$\}
   \State \textbf{return} $u \sim \pi^{*}_M(x)$
  \EndProcedure
  \State
  \Procedure{online-learning}{$x, u, y, r$}
	\State ``Update the posterior w.r.t. transition $<x, u, y, r>$''
  \EndProcedure
  \caption{BEB}\label{BEB_pseudocode}
  \end{algorithmic}
\end{algorithm}
\end{minipage}

\section{MDP distributions in detail}
In this section, we describe the MDPs drawn from the considered distributions in more detail. In addition, we also provide a formal description of the corresponding $\boldsymbol \theta$ (parameterising the FDM used to draw the transition matrix) and $\rho_M$ (the reward function).

\subsection{Generalised Chain distribution}
\label{appendix:gc}
On those MDPs, we can identify two possibly optimal behaviours:
\begin{itemize}
	\item	The agent tries to move along the chain, reaches the last state, and collect as many rewards as possible before returning to State $1$;
	\item	The agent gives up to reach State $5$ and tries to return to State $1$ as often as possible.
\end{itemize}

\subsubsection{Formal description}
{\centering $X = \{1, 2, 3, 4, 5\},\; U = \{1, 2, 3\}$\\}
\vspace{0.5cm}

\begin{minipage}{0.49\textwidth}
\centering
$\forall u \in U:$
\begin{align*}
	\theta^{GC}_{1, u} &= [1, 1, 0, 0, 0]\\
	\theta^{GC}_{2, u} &= [1, 0, 1, 0, 0]\\
	\theta^{GC}_{3, u} &= [1, 0, 0, 1, 0]\\
	\theta^{GC}_{4, u} &= [1, 0, 0, 0, 1]\\
	\theta^{GC}_{5, u} &= [1, 1, 0, 0, 1]
\end{align*}
\end{minipage}
\begin{minipage}{0.49\textwidth}
\centering
$\forall x, u \in X \times U:$
\begin{align*}
	\rho^{GC}(x, u, 1) &= 2.0\\
	\rho^{GC}(x, u, 5) &= 10.0\\
	\rho^{GC}(x, u, y) &= 0.0,\; \forall y \in X \setminus \{1, 5\}\\
\end{align*}
\end{minipage}

\subsection{Generalised Double-Loop distribution}
\label{appendix:gdl}
Similarly to the GC distribution, we can also identify two possibly optimal behaviours:
\begin{itemize}
	\item	The agent enters the ``good'' loop and tries to stay in it until the end;
	\item	The agent gives up and chooses to enter the ``bad'' loop as frequently as possible.
\end{itemize}

\subsubsection{Formal description}
{\centering $X = \{1, 2, 3, 4, 5, 6, 7, 8, 9\},\; U = \{1, 2\}$\\}
\vspace{0.5cm}

\begin{minipage}{0.49\textwidth}
\centering
$\forall u \in U:$
\begin{align*}
	\theta^{GDL}_{1, u} &= [0, 1, 0, 0, 0, 1, 0, 0, 0]\\
	\theta^{GDL}_{2, u} &= [0, 0, 1, 0, 0, 0, 0, 0, 0]\\
	\theta^{GDL}_{3, u} &= [0, 0, 0, 1, 0, 0, 0, 0, 0]\\
	\theta^{GDL}_{4, u} &= [0, 0, 0, 0, 1, 0, 0, 0, 0]\\
	\theta^{GDL}_{5, u} &= [1, 0, 0, 0, 0, 0, 0, 0, 0]\\
	\theta^{GDL}_{6, u} &= [1, 0, 0, 0, 0, 0, 1, 0, 0]\\
	\theta^{GDL}_{7, u} &= [1, 0, 0, 0, 0, 0, 0, 1, 0]\\
	\theta^{GDL}_{8, u} &= [1, 0, 0, 0, 0, 0, 0, 0, 1]\\
	\theta^{GDL}_{9, u} &= [1, 0, 0, 0, 0, 0, 0, 0, 0]
\end{align*}
\end{minipage}
\begin{minipage}{0.49\textwidth}
\centering
$\forall u \in U:$
\begin{align*}
	\rho^{GDL}(5, u, 1) &= 1.0\\
	\rho^{GDL}(9, u, 1) &= 2.0\\
	\rho^{GDL}(x, u, y) &= 0.0,\; \forall x \in X,\; \forall y \in X: y \neq 1
\end{align*}
\end{minipage}

\subsection{Grid distribution}
\label{appendix:grid}
MDPs drawn from the Grid distribution are 2-dimensional grids. Since the agents considered do not manage multi-dimensional state spaces, the following bijection was defined:
\[
	\{1, 2, 3, 4, 5\} \times \{1, 2, 3, 4, 5\} \rightarrow X = \{1, 2, \cdots, 25\}: n(i, j) = 5 (i - 1) + j
\]
where $i$ and $j$ are the row and column indexes of the cell on which the agent is.\\

When the agent reaches the \textbf{G} cell (in $(5, 5)$), it is directly moved to $(1, 1)$, and will perceive its reward of $10$. In consequence, State $(5, 5)$ is not reachable.\\

To move inside the Grid, the agent can perform four actions: $U = \{up, down, left, right\}$. Those actions only move the agent to one adjacent cell. However, each action has a certain probability to fail (depending on the cell on which the agent is). In case of failure, the agent does not move at all. Besides, if the agent tries to move out of the grid, it will not move either. Discovering a reliable (and short) path to reach the \textbf{G} cell will determine the success of the agent.

\subsubsection{Formal description}
{\centering $X = \{1, 2, \cdots, 25\}, U = \{up,\; down,\; left,\; right\}$\\}

\begin{alignat*}{3}
&\forall (i, j) &&\in \{1, 2, 3, 4, 5\} \times \{1, 2, 3, 4, 5\}\\
&\forall (k, l) &&\in \{1, 2, 3, 4, 5\} \times \{1, 2, 3, 4, 5\}:\\
\end{alignat*}
\vspace{-1.5cm}

\begin{minipage}{0.49\textwidth}
\centering
\begin{alignat*}{3}
	&\theta^{Grid}_{n(i, j), u}&&(n(i, j)) &&= 1,\; \forall u \in U\\
	\\
	&\theta^{Grid}_{n(i, j), up}&&(n(i - 1, j)) &&= 1, (i - 1) \geq 1\\
	&\theta^{Grid}_{n(i, j), down}&&(n(i + 1, j)) &&= 1, (i + 1) \leq 5, (i, j) \neq (4, 5)\\
	&\theta^{Grid}_{n(i, j), left}&&(n(i, j - 1)) &&= 1, (j - 1) \geq 1\\
	&\theta^{Grid}_{n(i, j), right}&&(n(i, j + 1)) &&= 1, (j + 1) \leq 5, (i, j) \neq (5, 4)\\
	\\
	&\theta^{Grid}_{n(4, 5), down}&&(n(1, 1)) &&= 1\\
	&\theta^{Grid}_{n(5, 4), right}&&(n(1, 1)) &&= 1\\
	\\
	&\theta^{Grid}_{n(i, j), u}&&(n(k, l)) &&= 0, else
\end{alignat*}
\end{minipage}
\begin{minipage}{0.49\textwidth}
\centering
\vspace{-1.5cm}
\begin{alignat*}{4}
	&\rho^{Grid}((4, 5), &&down, &&(1, 1)) &&= 10.0\\
	&\rho^{Grid}((5, 4), &&right, &&(1, 1)) &&= 10.0\\
	\\
	&\rho^{Grid}((i, j), &&u,&& (k, l)) &&= 0.0,\; \forall u \in U
\end{alignat*}
\end{minipage}

\section{Paired sampled $Z$-test}
\label{appendix:z-test}
Let $\pi_A$ and $\pi_B$ be the two agents we want to compare.
We played the two agents on the same $N$ MDPs, denoted by $M_1, \cdots, M_N$.
Let $R^{\pi_A}_{M_i}$ and $R^{\pi_B}_{M_i}$ be the scores we observed for the two agents on $M_i$.

\subsection*{Step 1 - Hypothesis}
We compute the mean and the standard deviation of the differences between the two sample sets, denoted by $\bar{x}_d$ and $\bar{s}_d$, respectively.
\begin{align*}
	\bar{x}_d &= \frac{1}{N} \sum_{i=1}^N   R^{\pi_A}_{M_i} - R^{\pi_B}_{M_i}\\
	\bar{s}_d  &= \frac{1}{N} \sum_{i=1}^N (\bar{x}_d - (R^{\pi_A}_{M_i} - R^{\pi_B}_{M_i}))^{2}
\end{align*}
If $N \geq 30$, $\bar{s}_d$ is a good estimation of $\sigma_d$, the standard deviation of the differences between the two populations ($\bar{s}_d \approx \sigma_d$).
In order words, $\sigma_d$ is the standard deviation we should observe when testing the two algorithms on a number of MDPs tending towards infinity.
This was always the case in our experiments.

We now set Hypothesis $H_0$ and Hypothesis $H_\alpha$:
\begin{align*}
	&H_0: \mu_d = 0\\
	&H_\alpha: \mu_d > 0
\end{align*}
Our goal is to determine if $\mu_d$, the mean of the differences between the two populations, is equal or greater than $0$.
More expressly, we want to know if the differences between the two agents' performances is significant ($H_\alpha$ is correct) or not ($H_0$ correct).
Only one of those hypotheses can be true.

\subsection*{Step 2 - Test statistic}
The test statistic consists to compute a certain value $Z$:
\[
	Z = \frac{\bar{x_d}}{\left. \sigma_d \middle/ \sqrt{N} \right.}
\]
This value will help us to determine if we should accept (or reject) hypothesis $H_\alpha$.

\subsection*{Step 3 - Rejection region}
Assuming we want our decision to be correct with a probability of failure of $\alpha$, we will have to compare $Z$ with $Z_\alpha$, a value of a Gaussian curve.
If $Z > Z_\alpha$, it means we are in the rejection region (R.R.) with a probability equal to $1 - \alpha$.
For a confidence of 95\%, $Z_\alpha$ should be equal to $1.645$.

\begin{center}
\includegraphics[width=0.35\textwidth]{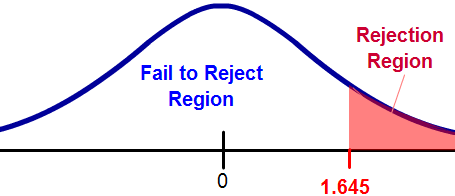}
\end{center}

Being in the R.R. means we have to reject Hypothesis $H_0$ (and accept Hypothesis $H_\alpha$).
In the order case, we have to accept Hypothesis $H_0$ (and reject Hypothesis $H_\alpha$).

\subsection*{Step 4 - Decision}
At this point, we have either accepted Hypothesis $H_0$ or Hypothesis $H_\alpha$.
\begin{itemize}
	\item \textbf{Accepting Hypothesis $\boldsymbol H_{\boldsymbol 0}$ ($\boldsymbol Z \boldsymbol < \boldsymbol Z_{\boldsymbol \alpha}$)}: The two algorithms $\pi_A$ and $\pi_B$ are not significantly different.
	\item \textbf{Accepting Hypothesis $\boldsymbol H_{\boldsymbol \alpha}$ ($\boldsymbol Z \boldsymbol \geq  \boldsymbol Z_{\boldsymbol \alpha}$)}: The two algorithms $\pi_A$ and $\pi_B$ are significantly different. Therefore, the algorithm with the greatest mean is definitely better with 95\% confidence.
\end{itemize}

\vskip 0.2in
\bibliography{biblio}

\begin{thebibliography}{14}
\providecommand{\natexlab}[1]{#1}
\providecommand{\url}[1]{\texttt{#1}}
\expandafter\ifx\csname urlstyle\endcsname\relax
  \providecommand{\doi}[1]{doi: #1}\else
  \providecommand{\doi}{doi: \begingroup \urlstyle{rm}\Url}\fi

\bibitem[Asmuth and Littman(2011)]{Asmuth11approachingbayes-optimalilty}
J.~Asmuth and M.~Littman.
\newblock Approaching {B}ayes-optimalilty using {M}onte-{C}arlo tree search.
\newblock In \emph{Proceedings of the 21st International Conference on
  Automated Planning and Scheduling}, 2011.

\bibitem[Asmuth et~al.(2009)Asmuth, Li, Littman, Nouri, and Wingate]{Asmuth09}
J.~Asmuth, L.~Li, M.L. Littman, A.~Nouri, and D.~Wingate.
\newblock A {B}ayesian sampling approach to exploration in {R}einforcement
  {L}earning.
\newblock In \emph{Proceedings of the Twenty-Fifth Conference on Uncertainty in
  Artificial Intelligence (UAI)}, pages 19--26. AUAI Press, 2009.

\bibitem[Audibert et~al.(2007)Audibert, Munos, and
  Szepesv{\'a}ri]{Audibert2007}
J.Y. Audibert, R.~Munos, and C.~Szepesv{\'a}ri.
\newblock Tuning bandit algorithms in stochastic environments.
\newblock In \emph{Algorithmic Learning Theory}, pages 150--165. Springer,
  2007.

\bibitem[Auer et~al.(2002)Auer, Cesa-Bianchi, and Fischer]{Auer2002}
P.~Auer, N.~Cesa-Bianchi, and P.~Fischer.
\newblock Finite-time analysis of the multiarmed bandit problem.
\newblock \emph{Machine learning}, 47\penalty0 (2):\penalty0 235--256, 2002.

\bibitem[Castro and Precup(2010)]{Castro10}
P.~S. Castro and D.~Precup.
\newblock Smarter sampling in model-based bayesian reinforcement learning.
\newblock In \emph{Machine Learning and Knowledge Discovery in Databases},
  pages 200--214. Springer, 2010.

\bibitem[Castronovo et~al.(2012)Castronovo, Maes, Fonteneau, and
  Ernst]{Castronovo12}
M.~Castronovo, F.~Maes, R.~Fonteneau, and D.~Ernst.
\newblock {L}earning exploration/exploitation strategies for single trajectory
  {R}einforcement {L}earning.
\newblock \emph{Journal of Machine Learning Research (JMLR)}, pages 1--9, 2012.

\bibitem[Castronovo et~al.(2014)Castronovo, Fonteneau, and Ernst]{Castronovo14}
M.~Castronovo, R.~Fonteneau, and D.~Ernst.
\newblock {B}ayes {A}daptive {R}einforcement {L}earning versus {O}ff-line
  {P}rior-based {P}olicy {S}earch: an {E}mpirical {C}omparison.
\newblock \emph{23rd annual machine learning conference of Belgium and the
  Netherlands (BENELEARN 2014)}, pages 1--9, 2014.

\bibitem[Dearden et~al.(1998)Dearden, Friedman, and Russell]{Dearden98}
R.~Dearden, N.~Friedman, and S.~Russell.
\newblock {B}ayesian {Q}-learning.
\newblock In \emph{Proceedings of Fifteenth National Conference on Artificial
  Intelligence (AAAI)}, pages 761--768. AAAI Press, 1998.

\bibitem[Dearden et~al.(1999)Dearden, Friedman, and Andre]{Dearden99}
R.~Dearden, N.~Friedman, and D.~Andre.
\newblock Model based {B}ayesian exploration.
\newblock In \emph{Proceedings of the Fifteenth Conference on Uncertainty in
  Artificial Intelligence (UAI)}, pages 150--159. Morgan Kaufmann, 1999.

\bibitem[Guez et~al.(2012)Guez, Silver, and Dayan]{Guez2012}
A.~Guez, D.~Silver, and P.~Dayan.
\newblock Efficient {B}ayes-adaptive {R}einforcement {L}earning using
  sample-based search.
\newblock In \emph{Neural Information Processing Systems (NIPS)}, 2012.

\bibitem[Kearns et~al.(2002)Kearns, Mansour, and Ng]{Kearns2002}
M.~Kearns, Y.~Mansour, and A.~Y. Ng.
\newblock A sparse sampling algorithm for near-optimal planning in large
  {M}arkov decision processes.
\newblock \emph{Machine Learning}, 49\penalty0 (2-3):\penalty0 193--208, 2002.

\bibitem[Kocsis and Szepesv{\'a}ri(2006)]{Kocsis2006}
L.~Kocsis and C.~Szepesv{\'a}ri.
\newblock Bandit based {M}onte-{C}arlo planning.
\newblock \emph{European Conference on Machine Learning (ECML)}, pages
  282--293, 2006.

\bibitem[Kolter and Ng(2009)]{Kolter09near-bayesianexploration}
J.~Zico Kolter and Andrew~Y. Ng.
\newblock {N}ear-{B}ayesian exploration in polynomial time.
\newblock In \emph{Proceedings of the 26th Annual International Conference on
  Machine Learning}, 2009.

\bibitem[Strens(2000)]{Strens00}
M.~Strens.
\newblock A {B}ayesian framework for {R}einforcement {L}earning.
\newblock In \emph{Proceedings of the Seventeenth International Conference on
  Machine Learning (ICML)}, pages 943--950. ICML, 2000.

\end{thebibliography}

\end{document}